\documentclass[11pt]{article}
\usepackage{acl}
\usepackage{times}
\usepackage{latexsym}
\usepackage[T1]{fontenc}
\usepackage[utf8]{inputenc}
\usepackage{microtype}
\usepackage{inconsolata}
\usepackage{graphicx}
\usepackage{amsmath}
\usepackage{booktabs}
\usepackage{tabularx}
\usepackage{cuted}
\usepackage{pgfplots}
\usepackage{xcolor}
\usepackage{listings}
\usepackage{url}
\newcommand{\toolid}[1]{\path{#1}}
\usepackage{makecell}

\newcommand{\yisen}[1]{\textcolor{black}{#1}}
\lstdefinestyle{promptstyle}{
  basicstyle=\ttfamily\scriptsize,
  breaklines=true,
  breakatwhitespace=false,
  columns=fullflexible,
  keepspaces=true,
  showstringspaces=false,
  frame=single
}
\pgfplotsset{compat=1.18}
\title{SING: Synthetic Intention Graph for Scalable Active Tool Discovery in LLM Agents}

\author{
\textsuperscript{*}\textbf{Qiao Xiao}\textsuperscript{1} \quad
\textsuperscript{*}\textbf{Haochen Shi}\textsuperscript{2} \quad
\textsuperscript{*}\textbf{Yisen Gao}\textsuperscript{2} \quad
\textbf{Wenbin Hu}\textsuperscript{2} \\
\textbf{Huihao Jing}\textsuperscript{2} \quad
\textbf{Tianshi Zheng}\textsuperscript{2} \quad
\textbf{Baixuan Xu}\textsuperscript{2} \quad
\textbf{Ziheng Zhang}\textsuperscript{3} \\
\textbf{Weiqi Wang}\textsuperscript{2} \quad
\textbf{Haoran Li}\textsuperscript{2} \quad
\textbf{Jiaxin Bai}\textsuperscript{4} \quad
\textsuperscript{\textdagger}\textbf{Yangqiu Song}\textsuperscript{2} \\
\textsuperscript{1}Cornell University \quad
\textsuperscript{2}The Hong Kong University of Science and Technology \\
\textsuperscript{3}The Ohio State University \quad
\textsuperscript{4}Hong Kong Baptist University
}

\begin{document}
\maketitle
\begingroup
% \renewcommand{\thefootnote}{}
% \footnotetext{\begin{tabular}[t]{@{}l@{\hspace{0.35em}}p{0.86\columnwidth}@{}}
% \textsuperscript{*} & These authors contributed equally to this work and share first authorship.\\
% \textsuperscript{\textdagger} & This author supervised the research and serves as the corresponding author.
% \end{tabular}}
\endgroup

\begin{abstract}
\yisen{Large language model (LLM) agents increasingly rely on agent harnesses that manage context, tools, and multi-turn execution, making tools a central interface for acting in realistic digital environments.
As harness-connected tool ecosystems expand to hundreds or thousands of APIs, services, and task-specific skills, exhaustive tool schema injection becomes costly and imposes a closed-world assumption that limits agents to a predefined static inventory.
Retrieval-augmented tool selection offers a natural alternative, but existing one-shot retrieval methods often fail to align isolated tool descriptions with the agent's true task intention, especially in long-horizon tasks where required capabilities emerge through decomposition, observations, and newly induced subgoals.
% Moreover, passive tool selection typically assumes a closed-world setting with a fixed tool inventory, making it incompatible with real-world agent execution environments in which available tools and affordances may be open-ended, dynamic, and discovered during interaction.
% To address this limitation, we propose \textbf{SING}, an intention-aware active tool discovery framework.
% SING introduces a hierarchical intention--tool graph that links abstract task intentions to concrete servers and tool capabilities, providing a structured discovery layer beyond surface-level description matching.
% During inference, SING dynamically aligns the agent's current request and observations with intention nodes to discover relevant tools, and then uses the matched intentions to select fine-grained tools within those servers.
% We build a unified corpus of 779 MCP servers and 7,471 tools, and evaluate SING on three real-world tool-use benchmarks.
% Experimental results show that SING improves both server discovery accuracy and downstream workflow success over embedding-based retrieval and active-discovery baselines, demonstrating the effectiveness of intention-aware graph discovery for large-scale agentic tool ecosystems.
We propose \textbf{SING}, an intention-aware active tool discovery framework that builds an intention--tool graph linking user intentions, tool capabilities, and tool collaboration patterns, and dynamically retrieves tools according to evolving task states.
Using a unified corpus of 7,471 tools, we evaluate SING on three real-world tool-use benchmarks.
SING improves Global Recall@5 by up to 59.8\% and downstream success rate by up to 28.9\% over baselines, while reducing full-corpus tool-schema exposure by 99.8\%, demonstrating that intention-aware graph structure enables more accurate and context-efficient tool
  discovery in large-scale agentic ecosystems.}
\end{abstract}

\section{Introduction}
%llm 不是独立部署-> tool愈发重要 tool已经成百上千的不断增加了,这种剧变让基于上下文注入的方法会失效,并且陷于close-world, 无法适应不断扩大的agent生态环境.使得agent需要从预定义的静态工具使用走向主动检索. 

\yisen{Large language models (LLMs) have achieved substantial progress in reasoning, planning, and natural language interaction~\citep{10.5555/3495724.3495883,cao2025largelanguagemodelsplanning}, but modern LLM agents are increasingly not deployed as standalone models. Instead, they are embedded in agent harnesses: external execution systems that organize task runs by managing context, tools, and control flow~\citep{HarnessEngineering}. These harnesses turn model outputs into concrete actions and feed observations back to the model, making tools a central interface through which agents operate in complex digital environments rather than merely answer natural-language queries. This shift is especially evident in recent open-world and computer-use settings, where benchmarks and runtimes such as ClawBench~\cite{zhang2026clawbenchaiagentscomplete}, OSWorld~\cite{xie2024osworld}, and WildClawBench~\cite{ding2026wildclawbench} evaluate agents on realistic tasks involving live websites, file systems, terminals, and external services. In such harness-based environments, tool ecosystems are no longer small fixed collections: they can grow to hundreds or thousands of APIs, services, and task-specific skills that agents may need to access during execution.}

A common tool-use paradigm is to inject all available tool schemas into the model context before execution~\citep{BFCL,tau-bench}. However, this context-injection strategy becomes increasingly ineffective as the tool ecosystem scales. Exposing hundreds or thousands of schemas consumes substantial context budget, introduces many irrelevant tools, and can weaken the model's retention of task-critical information~\citep{liu-etal-2024-lost}. More fundamentally, exhaustive schema injection imposes a closed-world assumption: the agent can only select from a predefined static inventory, rather than identifying, requesting, or experimenting with capabilities required by the evolving task state. As agent ecosystems continue to expand, tool use must therefore move beyond passive selection from a fixed context toward active discovery and retrieval of relevant tools during execution.

\begin{figure}[t]
\centering
\includegraphics[width=\linewidth]{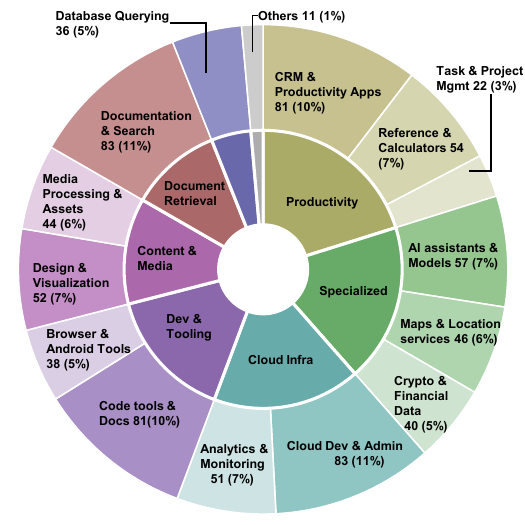}
\caption{Domain distribution of the MCP server corpus, consisting of 779 servers grouped into 15 domains.}
\label{fig:server_domain_distribution}
\end{figure}

% Retrieval-augmented tool selection enables agents to find tools in open-world ecosystems while keeping the context window manageable by exposing only a small subset of the full inventory~\citep{gan2025ragmcpmitigatingpromptbloat, tool2vec}.
% They usually adopt the one-shot design: selecting tools before the agent interacts with the environment, usually by matching the initial user utterance, or the model's immediate interpretation of it, against isolated tool descriptions.
% However, this design cannot align the ambiguous user intent, especially for real-world tasks requiring long-horizon multiturn interactions, where user intention may only become clear after task decomposition, intermediate observations, or newly induced subgoals.
% As a result, one-shot retrieval can favor tools aligned with the surface request while missing prerequisite, complementary, or downstream tools required for successful execution.
\yisen{A natural way to move beyond exhaustive tool injection is retrieval-augmented tool discovery: instead of exposing the full inventory, the agent retrieves a small set of candidate tools from a large tool library. This alleviates context explosion and enables models to interact with large real-world tool ecosystems~\citep{gan2025ragmcpmitigatingpromptbloat, tool2vec}. However, existing retrieval-based methods still struggle to align retrieved tools with true task intention. They typically follow a one-shot paradigm, retrieving tools before environment interaction by matching the initial user utterance, or the model's immediate interpretation of it, against isolated tool descriptions. This paradigm is fragile for real-world long-horizon tasks that require multi-turn interaction: user intent is often ambiguous, compositional, or only partially specified, while the required capabilities emerge through task decomposition, intermediate observations, and newly induced subgoals~\citep{wildtoolbench}. For example, a user may ask for the distance between two national parks, which superficially matches map or search tools, but successful execution may require geocoding both locations, extracting coordinates, and performing a numerical distance calculation. A retrieval system that only matches the surface request to isolated tool descriptions can miss these prerequisite and downstream capabilities.}

\yisen{To address these limitations, we propose \textbf{SING}, an intention-aware active tool discovery framework for scalable tool ecosystems. Motivated by prior work on user intent recognition and intention knowledge graphs~\citep{bai-etal-2026-intention, Bodonhelyi2024UserIR}, SING explicitly models tool discovery around task intentions rather than isolated tool descriptions. It builds an intention--tool graph that links abstract user intentions, concrete tool capabilities, and tool collaboration patterns, allowing agents to retrieve tools through structured intention matching and graph propagation. At inference time, SING uses a dynamic ReAct-style process to update discovery according to evolving task states and observations, so that newly induced subgoals can trigger additional tool retrieval instead of being constrained by a fixed one-shot candidate set.}

Our main contributions are summarized as follows:
\begin{itemize}
    \item We propose an active tool discovery framework that jointly models user intentions and tool structures to retrieve relevant servers and tools in large, heterogeneous tool spaces.
    \item We collect a large-scale MCP tool library containing 779 servers and 7,471 tools, and construct an intention--tool graph that captures tool capabilities and collaboration patterns.
    \item \yisen{We validate SING on MCP-Universe~\cite{mcpuniverse}, MCP-Atlas~\cite{bandi2026mcpatlaslargescalebenchmarktooluse}, and MCP-Bench~\cite{wang2025mcpbenchbenchmarkingtoolusingllm}, improving Global Recall@5 by up to 59.8\% and downstream success rate by up to 28.9\% while reducing full-corpus tool-schema exposure by 99.8\%.}

\end{itemize}

% 这个要是太占地方了可以改成粗体（1）。。。（2）。。。（3）。。。那样 ：）

\section{Related Work}
\subsection{Tool Augmented LLMs}

Tool use enables LLM agents to interact with external resources such as search engines, APIs, databases, and execution environments.
Early work such as ReAct \cite{react} establishes the basic reasoning--action paradigm, where models interleave reasoning, tool invocation, observation, and plan refinement.

To improve tool-use ability, supervised fine-tuning methods train models on annotated or synthetic tool-use demonstrations.
Representative work such as Toolformer \cite{schick2023toolformer}, Gorilla \cite{patil2024gorilla}, and ToolLLM \cite{qin2024toolllm} teaches models to imitate valid API calls and tool-use trajectories.
However, SFT is limited by the coverage and quality of offline demonstrations, and mainly encourages imitation rather than improvement through interaction.

Recent reinforcement-learning methods such as ToolRL \cite{qian2026toolrl}, ReTool \cite{feng2025retool}, and ToRL \cite{li2025torl} instead optimize tool use through execution feedback.
By learning from interaction outcomes, these methods better capture the agentic nature of tool use, improving when to invoke tools, which tools to select, and how to refine actions.

Despite this progress, existing tool-learning methods mostly assume that candidate tools or APIs are already available.
In realistic agentic ecosystems, tool use is increasingly open-world: agents must first discover relevant tools from a large, heterogeneous, and evolving tool space before deciding how to use them~\citep{Steinberger2026openclaw, ding2026wildclawbench}.
This motivates tool discovery in a scalable tool library.

\subsection{Tool Retrieval and Discovery}

Tool retrieval addresses the open-world step before tool use: identifying which tools should be exposed to an agent from a large, heterogeneous, and evolving ecosystem.
Most methods formulate this as an information retrieval problem, ranking tool descriptions or API documents by relevance to the user request.
Recent work improves this paradigm through better representations or retrieval structures: Tool2Vec \cite{tool2vec} learns usage-driven tool embeddingsz, AnyTool \cite{du2024anytool} performs hierarchical retrieval over large API collections, and ToolRerank \cite{zheng2024toolrerank} refines candidates with adaptive, hierarchy-aware reranking.

MCP-Zero \cite{fei2025mcpzeroactivetooldiscovery} moves beyond single-shot retrieval by enabling active tool discovery: agents can detect capability gaps, generate structured requests for missing tools, and discover tools through hierarchical routing. 
However, its routing still matches model-generated requests to natural-language descriptions, which are fragile on both sides. 
On the target side, MCP tool descriptions are often incomplete or ambiguous, with smells such as unclear purpose, missing usage guidance, unstated limitations, and opaque parameters~\citep{smelly}. 
On the query side, real-world tool-use interactions often contain ambiguous, underspecified, compositional, and cross-turn intentions that are difficult to compress into a single generated request~\citep{wildtoolbench}. 

SING addresses these limitations by formulating tool discovery as intention-aware graph retrieval.
It models structural relations among tasks, servers, and tools, enabling agents to discover not only individually relevant tools, but also collaborative tool structures needed for complex tasks.

\section{Methodology}

\begin{figure*}[t]
\centering
\includegraphics[width=0.98\textwidth]{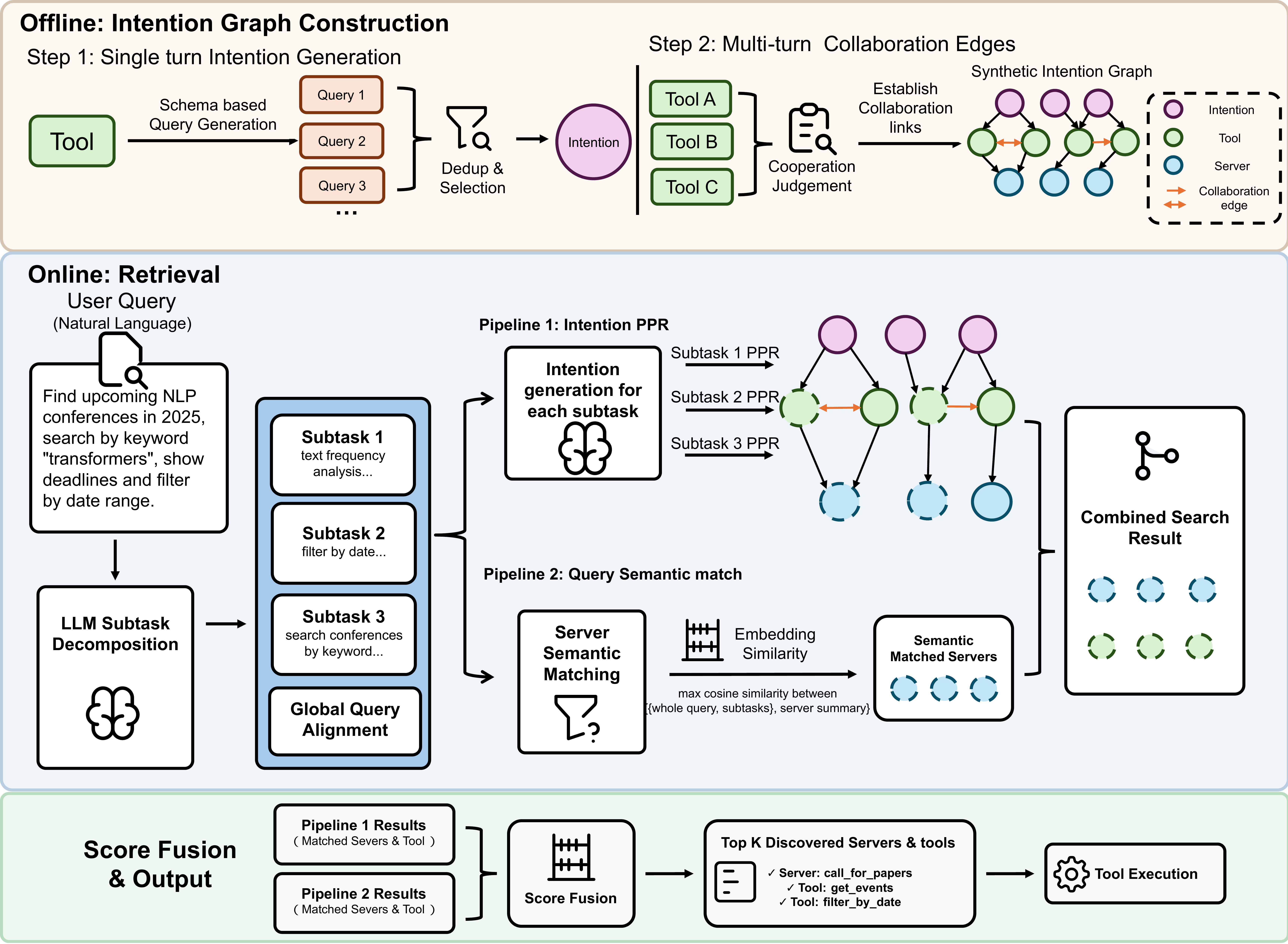}
\caption{Overview of SING.}
\label{fig:workflow_overview}
\vspace{-0.8em}
\end{figure*}

% \haochen{One important part is the difference between mcp-zero and our framework. I think one core advantage is we utilize intention graph to reveal furture tool calling, }
In this section, we present \textbf{SING}, a framework comprising three key components shown in Fig.~\ref{fig:workflow_overview}: (i) intention graph construction, (ii) a dynamic ReAct framework, and (iii) tool discovery layer.

\subsection{Intention Graph Construction}
\yisen{The intention graph is designed to expose the functional structure of a tool ecosystem, where tools are not isolated descriptions but are connected through shared goals and multi-step collaboration patterns. We use intentions as an intermediate abstraction between user requests and tool schemas because downstream tasks are usually expressed as goals, while tool descriptions often describe low-level operations or incomplete affordances. To obtain such intentions, we synthesize realistic tool-use queries and expand them into tool chains before extracting atomic verb-object intentions, rather than deriving intentions directly from schemas, which tends to produce generic and weakly task-aligned labels.}

\subsubsection{Intention Synthesis}\label{sec:query-intention-synthesis}

% To construct the intention graph, we first synthesize tool-use queries instead of directly deriving intentions from tool schemas.
% This design is motivated by the observation that schema-only generation tends to produce generic, low-diversity intentions that are weakly aligned with downstream user requests.
% By first generating realistic tool-use queries and expanding them into tool chains, we obtain richer task contexts from which more specific atomic intentions can be extracted.

\paragraph{Query synthesis.}

For each tool, we synthesize single tool queries from its parent server information, tool description, and schema.
We prompt an LLM to generate several natural language instructions together with their corresponding tool calls, including the target tool name andconcrete argument values.
To avoid trivial queries or direct paraphrases of tool descriptions, we filter too similar query by embedding similarity and low quality samples using LLM-as-a-judge.
\paragraph{Chain expansion.}

To generate more complex tasks, we expand single-tool queries into multi-tool chains that capture tool collaboration patterns.
Candidate next tools are selected using two signals.
First, we use an LLM to normalize tool inputs and outputs into unified type labels, and prioritize candidates with compatible input-output types.
Second, we ask an LLM to judge whether a candidate tool can naturally extend the current query and tool chain.
If valid, the LLM rewrites the query into a more complex multi-step task with the extended tool chain.

\phantomsection\label{par:atomic-intention-definition}

\paragraph{Atomic intention extraction.}
For each tool \(t\), we use an LLM to extract a small set of atomic intentions from its synthesized queries \(\mathcal{Q}_t\) and schema \(\sigma(t)\).
Each atomic intention is a concise verb-object phrase that describes a context-independent user goal supported by the tool.
We denote the normalized intention set of tool \(t\) as \(\mathcal{I}(t)\).
To reduce redundancy, semantically similar intentions are merged across tools and servers, producing shared intention nodes for future steps.
Appendix~\ref{app:query_intention_synthesis_example} and Table~\ref{tab:query_intention_chain_example} provide an example of query and intention synthesis for a GitHub tool chain.
\subsubsection{Graph Construction}

We construct a heterogeneous intention-tool graph \(G=(V,E)\), where \(V=V_s\cup V_t\cup V_i\) contains server, tool, and intention nodes.
We add two types of structural edges $E$.
For each tool \(t\) under server \(s\), we add a \texttt{has\_tool} edge from \(s\) to \(t\).
For each normalized intention \(i\) associated with \(t\), we add a \texttt{has\_intention} edge between \(t\) and \(i\).
Since intentions are globally normalized, tools from different servers can share intention nodes, enabling cross-server semantic connections beyond names, descriptions, or schemas.

Beyond the MCP hierarchy, we use synthesized multi-tool chains to capture tool collaboration.
For each chain, we add directed \texttt{tool\_next} edges between adjacent tools and undirected \texttt{tool\_cooccur} edges between tools appearing in the same chain.
Repeated relations are merged and assigned log-scaled frequency weights, \(w(t_a,t_b)=\log(1+c_{a,b})\), where \(c_{a,b}\) denotes how often the relation appears.
These weights are used by Personalized PageRank (PPR) in the discovery stage.

\subsection{Dynamic ReACT Framework}
We formulate SING as a dynamic ReAct-style framework that interleaves reasoning, tool discovery, tool invocation, and response generation.
Let \(x\) denote the user request, and let \(H_{j-1}\) and \(\mathcal{T}_{j-1}\) denote the interaction history and accessible tool set before step \(j\). The accessible tool set \(\mathcal{T}_{0}\) is initialized as empty.

At step \(j\), the agent samples a reasoning trace \(\tau_j\), an action type \(\rho_j\in\{\textsc{Discover},\textsc{Invoke},\textsc{Respond}\}\), and action content \(r_j\) given previous observations:
\begin{equation}
(\tau_j,\rho_j,r_j)
\sim
\pi_\theta
\bigl(
\cdot
\mid
x,H_{j-1},\mathrm{Desc}(\mathcal{C}_{j-1})
\bigr).
\end{equation}

When \(\rho_j=\textsc{Discover}\), the agent converts \(r_j\) into a state-conditioned query \(q^{(j)}=\phi_\theta(x,H_{j-1},r_j)\).
The discovery pipeline \(\mathcal{R}\), including server-level discovery and tool-level reranking, retrieves new tools \(\Delta\mathcal{T}_j=\mathcal{R}(q^{(j)})\).
The accessible tool set is updated as \(\mathcal{T}_j=\mathcal{T}_{j-1}\cup\Delta\mathcal{T}_j\), and the history is augmented with the discovery action and retrieved tool descriptions.
When \(\rho_j=\textsc{Invoke}\), the action content specifies a tool call \(r_j=(u_j,\xi_j)\), where \(u_j\in\mathcal{T}_{j-1}\).
The tool returns an observation \(o_j=u_j(\xi_j)\), and the history is updated as
\begin{equation}
H_j
=
H_{j-1}
\oplus
\bigl(
\tau_j,
\textsc{Invoke}(u_j,\xi_j),
o_j
\bigr).
\end{equation}
In this case, the accessible tool set remains unchanged, i.e., \(\mathcal{T}_j=\mathcal{T}_{j-1}\).
When \(\rho_j=\textsc{Respond}\), the agent terminates execution and returns \(r_j\) as the final response.

\begin{table*}[t]
\centering
\small
\setlength{\tabcolsep}{4pt}
\caption{MCP-Universe task-level evaluation results across task categories and overall performance. We report success rate (SR, \%) and average evaluator score (AE, \%). The best performance are \textbf{bold-faced}.}
\label{tab:mcpuniverse_task_evaluation}
\resizebox{\textwidth}{!}{%
\begin{tabular}{llcccccccc}
\toprule
\textbf{Setting} & \textbf{Method} 
& \shortstack{\textbf{Location}\\\textbf{Navigation}} 
& \shortstack{\textbf{Repository}\\\textbf{Management}} 
& \shortstack{\textbf{Financial}\\\textbf{Analysis}} 
& \shortstack{\textbf{3D}\\\textbf{Designing}} 
& \shortstack{\textbf{Browser}\\\textbf{Automation}} 
& \shortstack{\textbf{Web}\\\textbf{Searching}} 
& \multicolumn{2}{c}{\textbf{Overall}} \\
\cmidrule(lr){9-10}
 &  &  &  &  &  &  &  & \textbf{SR} & \textbf{AE} \\
\midrule
Default & Ground Truth Server 
& 31.4\% & 17.9\% & 60.0\% & 47.4\% & 38.2\% & 42.0\% 
& 40.3\% & 63.0\% \\
\midrule
Restricted & MCP-Zero 
& 22.9\% & 28.6\% & 60.0\% & \textbf{57.9\%} & 41.2\% & 44.0\% 
& 39.0\% & 51.2\% \\
Restricted & SING 
& \textbf{28.6\%} & \textbf{32.1\%} & \textbf{70.0\%} & 42.1\% & \textbf{50.0\%} & \textbf{50.0\%} 
& \textbf{47.1\%} & \textbf{69.5\%} \\
\midrule
Global & MCP-Zero 
& 28.6\% & 25.0\% & 62.5\% & 36.8\% & 35.3\% & 22.0\% 
& 35.0\% & 46.7\% \\
Global & SING 
& \textbf{31.4\%} & \textbf{26.4\%} & \textbf{65.0\%} & \textbf{52.6\%} & \textbf{52.9\%} & \textbf{46.0\%} 
& \textbf{45.1\%} & \textbf{72.7\%} \\
\bottomrule
\end{tabular}
}
\end{table*}

\begin{table*}[t]
\centering
\small
\setlength{\tabcolsep}{3pt}
\caption{Task-level evaluation results on MCP-Atlas and MCP-Bench. All values are reported as percentages. For MCP-Atlas, we report pass rate and mean coverage; for MCP-Bench, we report LLM-judge metrics and the overall score. \textbf{Bold-faced} denotes the best result within each discovery setting.}
\label{tab:combined_atlas_mcpbench}
\resizebox{\textwidth}{!}{%
\begin{tabular}{llccccccccc}
\toprule
\textbf{Setting} & \textbf{Method}
& \multicolumn{2}{c}{\textbf{MCP-Atlas}}
& \multicolumn{7}{c}{\textbf{MCP-Bench}} \\
\cmidrule(lr){3-4}\cmidrule(lr){5-11}
&
& \textbf{Pass Rate}
& \textbf{Mean Coverage}
& \multicolumn{2}{c}{\textbf{Task Completion}}
& \multicolumn{2}{c}{\textbf{Tool Usage}}
& \multicolumn{2}{c}{\textbf{Planning Effectiveness}}
& \textbf{Overall} \\
\cmidrule(lr){5-6}\cmidrule(lr){7-8}\cmidrule(lr){9-10}
&
& \textbf{(\%)}
& \textbf{(\%)}
& \shortstack{\textbf{Task}\\\textbf{Fulfillment}}
& \shortstack{\textbf{Information}\\\textbf{Grounding}}
& \shortstack{\textbf{Tool}\\\textbf{Appropriateness}}
& \shortstack{\textbf{Parameter}\\\textbf{Accuracy}}
& \shortstack{\textbf{Dependency}\\\textbf{Awareness}}
& \shortstack{\textbf{Parallelism and}\\\textbf{Efficiency}}
& \textbf{Score} \\
\midrule

Default
& Ground Truth Server
& 65.2\% & 71.0\%
& 53.5\% & 72.6\% & 74.6\% & 74.4\% & 62.3\% & 55.4\%
& 73.8\% \\

\midrule
Restricted
& MCP-Zero
& 45.4\% & 56.7\%
& 49.5\% & 62.9\% & 68.9\% & \textbf{68.3\%} & 58.7\% & \textbf{34.9\%}
& 67.6\% \\

Restricted
& SING
& \textbf{49.6\%} & \textbf{59.5\%}
& \textbf{50.1\%} & \textbf{66.8\%} & \textbf{71.4\%} & \textbf{68.3\%} & \textbf{60.1\%} & 33.4\%
& \textbf{68.4\%} \\

\midrule
Global
& MCP-Zero
& 34.0\% & 47.1\%
& 48.2\% & 61.1\% & \textbf{68.0\%} & 68.0\% & 57.7\% & 33.0\%
& 66.7\% \\

Global
& SING
& \textbf{40.2\%} & \textbf{51.0\%}
& \textbf{50.5\%} & \textbf{67.1\%} & 67.4\% & \textbf{69.0\%} & \textbf{60.8\%} & \textbf{35.3\%}
& \textbf{67.5\%} \\

\bottomrule
\end{tabular}
}
\end{table*}

\subsection{Discovery Layer}

Since tools are organized by MCP servers, SING adopts a hierarchical discovery pipeline that first retrieves relevant servers and then reranks the tools within them.

\subsubsection{Server-Level Discovery}
Given the user request \(x\), the agent first decomposes it into a global discovery request \(x_{\mathrm{global}}\) and a set of subtasks \(\{x_1,\ldots,x_n\}\) with the progress moves on.
The global request preserves the overall task intent, while the subtasks capture fine-grained functional needs.

Server-level discovery combines direct semantic matching with intention-guided graph propagation.
For direct matching, we score each server by its maximum embedding similarity to either \(x_{\mathrm{global}}\) or any subtask \(x_i\).
For graph propagation, we extract query-side intentions from the subtasks, match them to intention nodes in the graph, and use the matched intentions as PPR seeds.
We also seed tool nodes that are semantically relevant to the request, allowing evidence to propagate through server--tool--intention links and tool collaboration edges.
The resulting PPR score provides a graph-based relevance signal for each server.

The final server score is a weighted combination of normalized semantic and graph-based scores:
\begin{equation}
\mathrm{score}_{\mathrm{server}}(s)
=
\lambda_{\mathrm{sem}}\,\widehat{d}(s)
+
\lambda_{\mathrm{ppr}}\,\widehat{g}(s),
\end{equation}
where \(\widehat{d}(s)\) is the normalized direct semantic score computed by cosine similarity and \(\widehat{g}(s)\) is the normalized PPR score.
The top-\(K_s\) servers are then passed to tool-level discovery.

\subsubsection{Tool-Level Discovery}
Tool-level discovery reranks only the tools contained in the retrieved servers.
Let
\begin{equation}
\mathcal{T}
=
\bigcup_{s\in \mathcal{S}_{K_s}}\mathcal{T}(s),
\end{equation}
where \(\mathcal{S}_{K_s}\) denotes the top-\(K_s\) retrieved servers and \(\mathcal{T}(s)\) denotes the tools provided by server \(s\).

For each subtask \(x_i\), SING scores each candidate tool \(t\in\mathcal{T}\) using three complementary signals: tool-description matching, intention matching, and graph relevance:
\begin{equation}
\begin{aligned}
\mathrm{score}_{\mathrm{tool}}(t\mid x_i)
={}&
\lambda_{\mathrm{desc}}\,
\mathrm{sim}\!\left(x_i,\mathrm{desc}(t)\right) \\
&+
\lambda_{\mathrm{int}}\,
\max_{z\in\mathcal{I}(t)}
\mathrm{sim}\!\left(x_i,z\right) \\
&+
\lambda_{\mathrm{graph}}\,
\widehat{g}(t),
\end{aligned}
\end{equation}
where \(\mathrm{desc}(t)\) is the tool description, \(\mathcal{I}(t)\) is the set of intentions connected to \(t\), and \(\widehat{g}(t)\) is the normalized PPR score of the tool node.
For each subtask, we return the top-\(N\) tools, and the union of these tools forms the discovered tool set.

\section{Experiments}

\subsection{MCP Server Corpus}
\label{sec:server_corpus}
To evaluate SING in a realistic large-scale setting, we build a unified tool library from public MCP sources, including Glama, Smithery, and MCPHub, yielding 1,141 MCP servers.
To ensure evaluation stability and reproducibility, we filter out servers that cannot be reliably deployed in standalone settings, including deprecated or platform-managed servers, servers requiring private workspaces or user-specific credentials, and servers depending on non-text inputs.
For essential servers that require only standard API authentication, we manually configure the necessary tokens.
This process removes 31.7\% of the initial collection, resulting in a final corpus of 779 servers and 7,471 tools.
We further group the retained servers into 15 broad domains and report the distribution in Figure~\ref{fig:server_domain_distribution}. 
The comparison of diversity of our environments with other works is demonstrated in Appendix ~\ref{app:server_collection_details}.
%  and Figure~\ref{fig:cluster_tool_intention} visualizes the embedding space on tools. \haochen{shorter, we illustrate tool intention in experiments or discussion}

% \begin{figure}[t]
% \centering
% \includegraphics[width=\linewidth]{servers.pdf}
% \caption{Domain distribution of the final MCP server corpus after filtering. The corpus contains 779 servers grouped into 15 domains via HDBSCAN clustering.}
% \label{fig:server_domain_distribution}
% \end{figure}

\subsection{Experimental Setup and Metrics}

To construct the tool discovery layer, we use the MCP server corpus described in Section~\ref{sec:server_corpus} as the retrieval corpus and employ DeepSeek-V3.2 \cite{deepseek-v3} for intention synthesis and graph construction.
This process produces an intention--tool graph with 31,756 nodes and 57,524 edges, which serves as the retrieval backbone.
Unless otherwise specified, we use Qwen3-Embedding-8B \cite{qwenembedding} as the embedding extractor and set all retrieval weights to 1.

We evaluate SING under two retrieval settings.
The \textbf{Global} setting retrieves from the full MCP corpus, testing open-world discovery in a large and noisy tool ecosystem.
The \textbf{Restricted} setting retrieves only from servers associated with the corresponding benchmark, reducing the search space while retaining non-trivial distractors.
We assess server discovery using Recall@5, Full-Recall@5, and Mean Reciprocal Rank (MRR).
Recall@5 measures whether at least one gold server appears in the retrieved results, whereas Full-Recall@5 requires all gold servers to be retrieved.
MRR evaluates ranking quality by the rank of the first relevant server, computed as \(\mathrm{MRR}=\frac{1}{N}\sum_{i=1}^{N}1/\mathrm{rank}_i\), where \(\mathrm{rank}_i\) denotes the rank of the first relevant server for example \(i\).
Together, these metrics capture partial coverage, complete target coverage, and ranking quality.

Beyond retrieval quality, we also evaluate downstream task execution.
Unlike conventional tool-retrieval settings that assume a fixed gold tool subset, tools discovered in our setting are executable, and non-gold tools may still solve the task when they provide equivalent or complementary functionality.
We therefore evaluate SING on three complementary MCP benchmarks: MCP-Universe~\citep{mcpuniverse}, MCP-Atlas~\citep{bandi2026mcpatlaslargescalebenchmarktooluse}, and MCP-Bench~\citep{wang2025mcpbenchbenchmarkingtoolusingllm}.
MCP-Universe focuses on domain-specific tasks with execution-based evaluation, MCP-Atlas covers broad multi-step tasks over real servers with claims-based scoring, and MCP-Bench emphasizes fuzzy, dependency-driven tool orchestration using both rule-based and LLM-judge evaluation.
Unless otherwise specified, we follow the original evaluation protocols of each benchmark.
We use DeepSeek-V4-Pro~\citep{deepseek2026v4} as the runner model and Kimi-K2.5~\citep{kimiteam2026kimik25visualagentic} as the judge model for benchmarks requiring LLM-based evaluation.
As baselines, we compare SING with embedding-based one-shot retrieval and MCP-Zero, a dynamic ReAct-style active discovery framework.
Additional benchmark details are provided in Appendix~\ref{appendix:benchmark_details}.
\subsection{Server Discovery Accuracy}

\begin{table}[ht]
\centering
\small
\setlength{\tabcolsep}{4pt}
\caption{Server discovery performance across three MCP benchmarks under Global and Restricted retrieval settings.}
\label{tab:server_retrieval_results}
\resizebox{\linewidth}{!}{%
\begin{tabular}{llccc}
\toprule
\textbf{Dataset} & \textbf{Method} & \textbf{Recall@5} & \shortstack{\textbf{Full-Recall@5}} & \textbf{MRR} \\
\midrule
\multicolumn{5}{l}{\textbf{Global setting}} \\
\addlinespace

MCPBench & Embedding Only & 0.7436 & 0.5673 & 0.7798 \\
MCPBench & MCP-Zero & 0.7901 & 0.6250 & 0.8631 \\
% MCPBench & SING w/o Intention & 0.8974 & 0.7600 & 0.9038 \\
MCPBench & SING & \textbf{0.9022} & \textbf{0.7692} & \textbf{0.9199} \\

\addlinespace
MCP-Universe & Embedding Only & 0.3111 & 0.2457 & 0.1797 \\
MCP-Universe & MCP-Zero & 0.4339 & 0.3664 & 0.2948 \\
% MCP-Universe & SING w/o Intention & 0.5136 & 0.3534 & 0.4744 \\
MCP-Universe & SING & \textbf{0.5438} & \textbf{0.3750} & \textbf{0.5089} \\

\addlinespace
MCP-Atlas & Embedding Only & 0.1612 & 0.0200 & 0.2284 \\
MCP-Atlas & MCP-Zero & 0.1985 & 0.0260 & 0.2647 \\
% MCP-Atlas & SING w/o Intention & 0.3100 & \textbf{0.0540} & 0.4330 \\
MCP-Atlas & SING & \textbf{0.3172} & \textbf{0.0540} & \textbf{0.4362} \\

\midrule
\multicolumn{5}{l}{\textbf{Restricted setting}} \\
\addlinespace

MCPBench & Embedding Only & 0.9471 & 0.8846 & 0.9856 \\
MCPBench & MCP-Zero & 0.9487 & 0.8750 & 0.9952 \\
% MCPBench & SING w/o Intention & \textbf{0.9936} & \textbf{0.9808} & \textbf{1.0000} \\
MCPBench & SING & \textbf{0.9936} & \textbf{0.9808} & \textbf{1.0000} \\

\addlinespace
MCP-Universe & Embedding Only & 0.7974 & 0.6983 & 0.7071 \\
MCP-Universe & MCP-Zero & 0.8606 & 0.7284 & 0.8691 \\
% MCP-Universe & SING w/o Intention & \textbf{0.9116} & \textbf{0.8233} & 0.9315 \\
MCP-Universe & SING & \textbf{0.9116} & \textbf{0.8233} & \textbf{0.9348} \\

\addlinespace
MCP-Atlas & Embedding Only & 0.4646 & 0.1380 & 0.6505 \\
MCP-Atlas & MCP-Zero & 0.5263 & 0.1820 & 0.6917 \\
% MCP-Atlas & SING w/o Intention & 0.5799 & 0.2560 & 0.7451 \\
MCP-Atlas & SING & \textbf{0.5897} & \textbf{0.2680} & \textbf{0.7456} \\

\bottomrule
\end{tabular}
}
\end{table}

\yisen{Table~\ref{tab:server_retrieval_results} reports server discovery accuracy under the Global and Restricted settings. \textsc{Embedding Only} denotes one-shot retrieval with Qwen3-Embedding-8B, while \textsc{MCP-Zero} is the active-discovery baseline that decomposes each task into subtasks and retrieves top-5 servers.}

\yisen{SING achieves the best overall performance across all three benchmarks. The gains are most pronounced in the Global setting, where the server pool is large and noisy: compared with \textsc{MCP-Zero}, SING improves Recall@5 by 0.112, 0.110, and 0.119 on MCP-Bench, MCP-Universe, and MCP-Atlas, respectively, corresponding to relative gains of 14.2\%, 25.3\%, and 59.8\%. It also improves MRR by 0.057, 0.214, and 0.172 on the three benchmarks.}

\yisen{SING remains effective in the Restricted setting, improving Recall@5 over \textsc{MCP-Zero} by 0.045, 0.051, and 0.063 on MCP-Bench, MCP-Universe, and MCP-Atlas, respectively. These results indicate that intention-aware graph discovery helps both in open-world retrieval and in selecting among already relevant candidate servers. Results with different decomposition models are provided in Table~\ref{tab:decompose_model_benchmark}.}
\subsection{Downstream Task Evaluation}
Because non-ground-truth tools may also provide equivalent or complementary functionality, we further evaluate SING in executable task environments.
Table~\ref{tab:mcpuniverse_task_evaluation} reports the MCP-Universe results.
\yisen{SING consistently outperforms MCP-Zero under both discovery settings, improving the overall success rate by 8.1 and 10.1 percentage points in the Restricted and Global settings, respectively, corresponding to relative gains of 20.8\% and 28.9\%.}
Notably, SING also surpasses the \textsc{Default} setting, where only the ground-truth servers are provided, by 6.8 and 4.8 percentage points in overall success rate under Restricted and Global discovery.
This suggests that intention-guided retrieval does more than identify relevant servers: by constructing more focused, subtask-specific tool contexts, it can also reduce irrelevant tool noise and facilitate downstream inference.

Table~\ref{tab:combined_atlas_mcpbench} reports results on MCP-Atlas and MCP-Bench.
On MCP-Atlas, SING improves over MCP-Zero in both pass rate and mean coverage under both Restricted and Global settings, with particularly clear gains in the more challenging Global setting.
Its absolute performance remains below \textsc{Default}, which is consistent with the lower retrieval accuracy observed for MCP-Atlas.
On MCP-Bench, SING also achieves higher overall scores than MCP-Zero under both settings, indicating stronger end-to-end task execution despite the benchmark's fuzzy and dependency-driven tool requirements.
Together, these results show that SING's intention-aware discovery improves not only retrieval accuracy, but also practical tool-use performance across diverse execution-based benchmarks.
\subsection{Tool Schema Exposure Cost}

\begin{figure}[t]
\centering
\includegraphics[width=\linewidth]{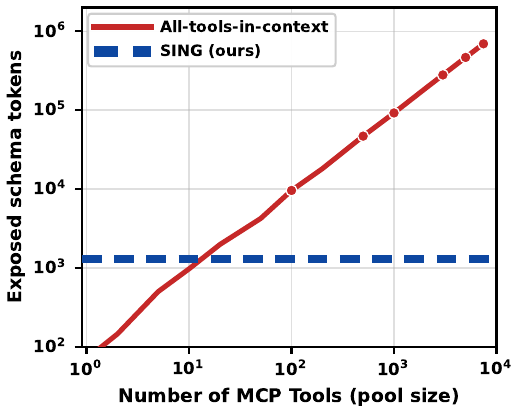}
\caption{Schema-token exposure under \textsc{All-tools-in-context} and SING retrieval as the MCP tool pool size increases.}
\label{fig:token_cost}
\end{figure}

\begin{figure*}[h]
\centering
\includegraphics[width=0.98\textwidth]{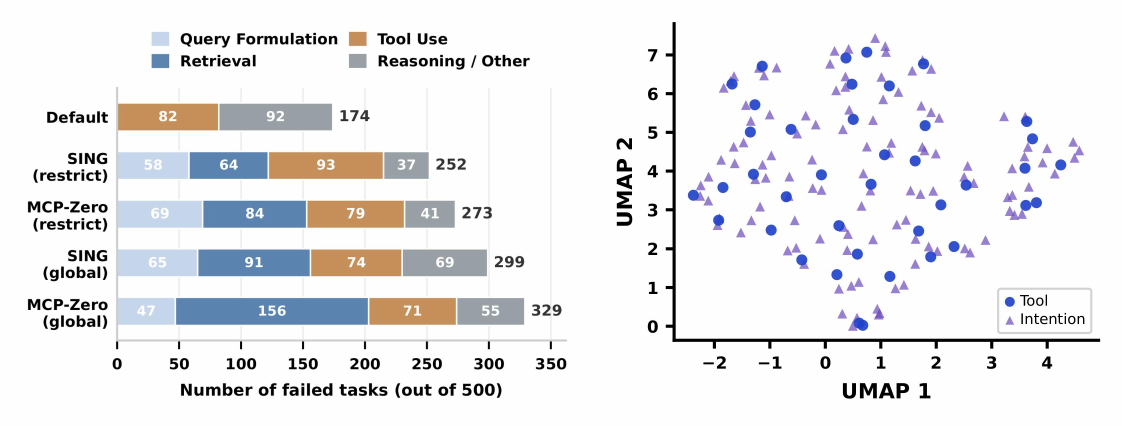}
\caption{Analysis of SING discovery behavior. Left: failure analysis on MCP-Atlas under Restricted and Global retrieval settings. Right: visualization of a sample of 40 tools and 120 intentions with 2D UMAP.}
\label{fig:analysis_combined}
\end{figure*}

\yisen{To assess the token efficiency of SING, we measure the average number of exposed tool-schema tokens across the three benchmarks under the Global setting as the tool pool size increases. As shown in Figure~\ref{fig:token_cost}, the context cost of \textsc{All-tools-in-context} grows rapidly with the size of the tool library, since every additional tool introduces more schema text into the prompt. In contrast, SING keeps the exposed schema cost nearly constant by retrieving only a small task-relevant subset, making its context overhead largely independent of the full library size. In the full-corpus setting, SING reduces schema exposure from 693,574 to 1,298 tokens on average, a 99.8\% reduction.}

\subsection{Analysis and Discussion}

\paragraph{Error analysis.}
We conduct failure analysis on MCP-Atlas because it contains broad multi-step tasks over diverse MCP servers and has a lower success rate than the other benchmarks, making it useful for studying how discovery errors affect downstream execution.
For each failed trajectory, we prompt an LLM judge to inspect the execution trace, including the task, discovery queries, retrieved tools, tool calls, outputs, and final answer, and assign one primary failure source.
\textbf{Query Formulation} covers failures where the agent fails to express the correct tool need during planning or decomposition.
\textbf{Retrieval} covers cases where the query is valid but discovery fails to return a tool with the required capability.
\textbf{Tool Use} refers to incorrect use of an available tool, such as wrong arguments, malformed calls, incorrect call order, or output misinterpretation.
\textbf{Reasoning / Other} includes failures after the required tools are available and used, such as incomplete reasoning, premature stopping, forgotten constraints, or execution issues.
Detailed definitions, prompts, and examples are provided in Appendices~\ref{app:failure_analysis_taxonomy} and~\ref{app:error_examples}.

As shown in Figure~\ref{fig:analysis_combined} left, SING reduces failed tasks from 273 to 252 in the Restricted setting and from 329 to 299 in the Global setting.
The main gains come from fewer retrieval errors: compared with MCP-Zero, SING reduces such failures from 84 to 64 under Restricted discovery and from 156 to 91 under Global discovery.
This indicates that intention-aware graph retrieval is especially useful in large, noisy MCP libraries.
Meanwhile, some execution-stage errors remain or increase slightly because improved retrieval allows more trajectories to reach later stages, exposing bottlenecks in parameter filling, output parsing, tool sequencing, and final reasoning.
Overall, SING shifts the bottleneck from upstream discovery to downstream agent execution.

% \paragraph{Why intentions help discovery.}
% Intention nodes complement raw tool descriptions with goal-level semantics.
% Figure~\ref{fig:cluster_tool_intention} shows that tool and intention embeddings are locally aligned, suggesting that synthesized intentions capture functional semantics close to tool usage.
% This abstraction better matches subtask queries, which often describe user goals rather than schema wording.
% Moreover, shared or similar intentions connect related tools, including across servers, creating effective PPR propagation paths.
% Thus, discovery benefits from higher-level functional structure beyond direct semantic matching.
\paragraph{Why intentions help discovery.}
Intention nodes enrich raw tool descriptions with goal-level semantics.
As shown in Figure~\ref{fig:analysis_combined} right, tool and intention embeddings are locally aligned, indicating that synthesized intentions capture functional meanings closely related to tool usage.
This abstraction better matches subtask queries, which are typically expressed as user goals rather than schema-level descriptions.
In addition, shared or similar intentions connect functionally related tools, including tools across different servers, enabling effective propagation through PPR.
Thus, intention-aware discovery benefits from higher-level functional structure beyond direct semantic matching.

\section{Conclusion}

% This paper introduces \textbf{SING}, an intention-aware graph framework for scalable tool discovery in real-world MCP ecosystems.
% SING addresses the limitations of existing tool selection methods by moving beyond full schema exposure and independent text-based tool matching.
% By constructing a synthetic intention--tool graph that captures both tool capabilities and collaboration patterns, SING provides structured semantic signals that enable agents to actively discover relevant servers and tools in large, heterogeneous tool spaces.
% Experiments on MCP-Universe, MCP-Atlas, and MCP-Bench show that SING consistently improves server discovery accuracy and downstream workflow performance under both restricted and global discovery settings.
% Further analysis shows that these gains mainly come from reducing upstream discovery failures, suggesting that SING mitigates a critical bottleneck in multi-step tool-use agents.
\yisen{This paper introduces \textbf{SING}, an intention-aware graph framework for scalable tool discovery in large tool ecosystems. SING constructs an intention--tool graph that links user intentions and tools, enabling agents to retrieve tools through structured intention matching and graph propagation. Experiments on three benchmarks show that SING improves server discovery accuracy and downstream task performance while substantially reducing tool-schema exposure. These results suggest that our method offers a practical path toward more accurate, context-efficient, and adaptive tool use in large-scale agentic ecosystems.}

\newpage

\section*{Limitations}
While SING improves tool discovery in large MCP ecosystems and multi-step tasks, several limitations remain.

\paragraph{Evaluation Cost.}
Downstream MCP evaluation is costly because it requires multi-step agent interactions, real tool calls, and LLM-based judging.
Therefore, our experiments focus on a practical evaluation setting with strong cost-effective models.
Future work can further examine SING across broader model families and agent backbones.

% \paragraph{Server Coverage and Reproducibility.}
% Our server corpus focuses on MCP servers that can be reliably collected, deployed, and evaluated in a reproducible text-based environment.
% For existing benchmarks, we follow their original server settings and evaluation protocols as closely as possible.
% For newly collected servers, we exclude instances that cannot be consistently reproduced.
% This design improves evaluation reliability, though future work can extend SING to more diverse server environments, including private, multimodal, or rapidly changing services.

\paragraph{Execution Variability.}
Real-world MCP execution is less controlled than static retrieval evaluation.
Tool outputs may vary due to live service updates, availability, or changing external content, and downstream performance can also depend on the agent's generated tool arguments.
SING reduces upstream discovery failures, but it does not remove all variability inherent to real-world tool-use environments.

\section*{Ethical Considerations}

This work complies with the ACL Code of Ethics.
All MCP servers used in our corpus are collected from publicly available MCP repositories or marketplaces, and all downstream evaluations are conducted on public MCP benchmarks.
We do not collect or release non public user data, private workspaces, API keys, access tokens, or user credentials.
For servers that require authentication, credentials are used only for local evaluation and are excluded from all released artifacts.
When using public MCP servers and external APIs, we follow the corresponding terms of service.

We used LLMs to assist with grammar polishing, wording refinement, LaTeX editing, and limited code debugging.
They were not used to generate, design the core research ideas, or draw scientific conclusions.
All LLM assisted outputs were manually reviewed and revised by the authors.

\bibliography{custom}

@misc{fei2025mcpzeroactivetooldiscovery,
      title={MCP-Zero: Active Tool Discovery for Autonomous LLM Agents}, 
      author={Xiang Fei and Xiawu Zheng and Hao Feng},
      year={2025},
      eprint={2506.01056},
      archivePrefix={arXiv},
      primaryClass={cs.AI},
      url={https://arxiv.org/abs/2506.01056}, 
}

@article{Bodonhelyi2024UserIR,
  title={User Intent Recognition and Satisfaction with Large Language Models: A User Study with ChatGPT},
  author={Anna Bodonhelyi and Efe Bozkir and Shuo Yang and Enkelejda Kasneci and Gjergji Kasneci},
  journal={ArXiv},
  year={2024},
  volume={abs/2402.02136},
  url={https://api.semanticscholar.org/CorpusID:267412438}
}

@misc{xu2025toucansynthesizing15mtoolagentic,
      title={TOUCAN: Synthesizing 1.5M Tool-Agentic Data from Real-World MCP Environments}, 
      author={Zhangchen Xu and Adriana Meza Soria and Shawn Tan and Anurag Roy and Ashish Sunil Agrawal and Radha Poovendran and Rameswar Panda},
      year={2025},
      eprint={2510.01179},
      archivePrefix={arXiv},
      primaryClass={cs.LG},
      url={https://arxiv.org/abs/2510.01179}, 
}

@misc{mcpuniverse,
  title={MCP-Universe: Benchmarking Large Language Models with Real-World Model Context Protocol Servers},
  author={Ziyang Luo and Zhiqi Shen and Wenzhuo Yang and Zirui Zhao and Prathyusha Jwalapuram and Amrita Saha and Doyen Sahoo and Silvio Savarese and Caiming Xiong and Junnan Li},
  year={2025},
  eprint={2508.14704},
  archivePrefix={arXiv},
  primaryClass={cs.AI},
  url={https://arxiv.org/abs/2508.14704}, 
}

@misc{chen2026divescalingdiversityagentic,
      title={DIVE: Scaling Diversity in Agentic Task Synthesis for Generalizable Tool Use}, 
      author={Aili Chen and Chi Zhang and Junteng Liu and Jiangjie Chen and Chengyu Du and Yunji Li and Ming Zhong and Qin Wang and Zhengmao Zhu and Jiayuan Song and Ke Ji and Junxian He and Pengyu Zhao and Yanghua Xiao},
      year={2026},
      eprint={2603.11076},
      archivePrefix={arXiv},
      primaryClass={cs.AI},
      url={https://arxiv.org/abs/2603.11076}, 
}

@misc{wang2025mcpbenchbenchmarkingtoolusingllm,
      title={MCP-Bench: Benchmarking Tool-Using LLM Agents with Complex Real-World Tasks via MCP Servers}, 
      author={Zhenting Wang and Qi Chang and Hemani Patel and Shashank Biju and Cheng-En Wu and Quan Liu and Aolin Ding and Alireza Rezazadeh and Ankit Shah and Yujia Bao and Eugene Siow},
      year={2025},
      eprint={2508.20453},
      archivePrefix={arXiv},
      primaryClass={cs.CL},
      url={https://arxiv.org/abs/2508.20453}, 
}

@misc{bandi2026mcpatlaslargescalebenchmarktooluse,
      title={MCP-Atlas: A Large-Scale Benchmark for Tool-Use Competency with Real MCP Servers}, 
      author={Chaithanya Bandi and Ben Hertzberg and Geobio Boo and Tejas Polakam and Jeff Da and Sami Hassaan and Manasi Sharma and Andrew Park and Ernesto Hernandez and Dan Rambado and Ivan Salazar and Rafael Cruz and Chetan Rane and Ben Levin and Brad Kenstler and Bing Liu},
      year={2026},
      eprint={2602.00933},
      archivePrefix={arXiv},
      primaryClass={cs.SE},
      url={https://arxiv.org/abs/2602.00933}, 
}

@inproceedings{bai-etal-2026-intention,
    title = "Intention Knowledge Graph Construction for User Intention Relation Modeling",
    author = "Bai, Jiaxin  and
      Wang, Zhaobo  and
      Cheng, Junfei  and
      Yu, Dan  and
      Huang, Zerui  and
      Wang, Weiqi  and
      Liu, Xin  and
      Luo, Chen  and
      Zhu, Yanming  and
      Li, Bo  and
      Song, Yangqiu",
    editor = "Demberg, Vera  and
      Inui, Kentaro  and
      Marquez, Llu{\'i}s",
    booktitle = "Proceedings of the 19th Conference of the {E}uropean Chapter of the {A}ssociation for {C}omputational {L}inguistics (Volume 1: Long Papers)",
    month = mar,
    year = "2026",
    address = "Rabat, Morocco",
    publisher = "Association for Computational Linguistics",
    url = "https://aclanthology.org/2026.eacl-long.21/",
    doi = "10.18653/v1/2026.eacl-long.21",
    pages = "466--484",
    ISBN = "979-8-89176-380-7"
}

@inproceedings{
  wildtoolbench,
  title={Benchmarking {LLM} Tool-Use in the Wild},
  author={Peijie Yu and Wei Liu and Yifan Yang and Jinjian Li and Zelong Zhang and Xiao Feng and Feng Zhang},
  booktitle={The Fourteenth International Conference on Learning Representations},
  year={2026},
  url={https://openreview.net/forum?id=yz7fL5vfpn}
}

@inproceedings{10.5555/3495724.3495883,
author = {Brown, Tom B. and Mann, Benjamin and Ryder, Nick and Subbiah, Melanie and Kaplan, Jared and Dhariwal, Prafulla and Neelakantan, Arvind and Shyam, Pranav and Sastry, Girish and Askell, Amanda and Agarwal, Sandhini and Herbert-Voss, Ariel and Krueger, Gretchen and Henighan, Tom and Child, Rewon and Ramesh, Aditya and Ziegler, Daniel M. and Wu, Jeffrey and Winter, Clemens and Hesse, Christopher and Chen, Mark and Sigler, Eric and Litwin, Mateusz and Gray, Scott and Chess, Benjamin and Clark, Jack and Berner, Christopher and McCandlish, Sam and Radford, Alec and Sutskever, Ilya and Amodei, Dario},
title = {Language models are few-shot learners},
year = {2020},
isbn = {9781713829546},
publisher = {Curran Associates Inc.},
address = {Red Hook, NY, USA},
booktitle = {Proceedings of the 34th International Conference on Neural Information Processing Systems},
articleno = {159},
numpages = {25},
url = {https://proceedings.neurips.cc/paper/2020/hash/1457c0d6bfcb4967418bfb8ac142f64a-Abstract.html},
location = {Vancouver, BC, Canada},
series = {NIPS '20}
}

@misc{cao2025largelanguagemodelsplanning,
      title={Large Language Models for Planning: A Comprehensive and Systematic Survey}, 
      author={Pengfei Cao and Tianyi Men and Wencan Liu and Jingwen Zhang and Xuzhao Li and Xixun Lin and Dianbo Sui and Yanan Cao and Kang Liu and Jun Zhao},
      year={2025},
      eprint={2505.19683},
      archivePrefix={arXiv},
      primaryClass={cs.AI},
      url={https://arxiv.org/abs/2505.19683}, 
}

@article{liu-etal-2024-lost,
    title = "Lost in the Middle: How Language Models Use Long Contexts",
    author = "Liu, Nelson F.  and
      Lin, Kevin  and
      Hewitt, John  and
      Paranjape, Ashwin  and
      Bevilacqua, Michele  and
      Petroni, Fabio  and
      Liang, Percy",
    journal = "Transactions of the Association for Computational Linguistics",
    volume = "12",
    year = "2024",
    address = "Cambridge, MA",
    publisher = "MIT Press",
    url = "https://aclanthology.org/2024.tacl-1.9/",
    doi = "10.1162/tacl_a_00638",
    pages = "157--173"
}

@misc{mo2026livemcpbenchagentsnavigateocean,
      title={LiveMCPBench: Can Agents Navigate an Ocean of MCP Tools?}, 
      author={Guozhao Mo and Wenliang Zhong and Jiawei Chen and Qianhao Yuan and Xuanang Chen and Yaojie Lu and Hongyu Lin and Ben He and Xianpei Han and Le Sun},
      year={2026},
      eprint={2508.01780},
      archivePrefix={arXiv},
      primaryClass={cs.AI},
      url={https://arxiv.org/abs/2508.01780}, 
}

@misc{wang2026mcpflowfacilitatingllmagents,
      title={MCP-Flow: Facilitating LLM Agents to Master Real-World, Diverse and Scaling MCP Tools}, 
      author={Wenhao Wang and Peizhi Niu and Zhao Xu and Zhaoyu Chen and Jian Du and Yaxin Du and Xianghe Pang and Keduan Huang and Yanfeng Wang and Qiang Yan and Siheng Chen},
      year={2026},
      eprint={2510.24284},
      archivePrefix={arXiv},
      primaryClass={cs.AI},
      url={https://arxiv.org/abs/2510.24284}, 
}

@misc{liu2026mcpagentbenchrealworldtaskbenchmark,
      title={MCPAgentBench: A Real-world Task Benchmark for Evaluating LLM Agent MCP Tool Use}, 
      author={Wenrui Liu and Zixiang Liu and Elsie Dai and Wenhan Yu and Lei Yu and Tong Yang and Jinjun Han and Hong Gao},
      year={2026},
      eprint={2512.24565},
      archivePrefix={arXiv},
      primaryClass={cs.AI},
      url={https://arxiv.org/abs/2512.24565}, 
}

@misc{fan2025mcptoolbenchlargescaleai,
      title={MCPToolBench++: A Large Scale AI Agent Model Context Protocol MCP Tool Use Benchmark}, 
      author={Shiqing Fan and Xichen Ding and Liang Zhang and Linjian Mo},
      year={2025},
      eprint={2508.07575},
      archivePrefix={arXiv},
      primaryClass={cs.AI},
      url={https://arxiv.org/abs/2508.07575}, 
}

@misc{deepseek2026v4,
  title        = {DeepSeek-V4: Towards Highly Efficient Million-Token Context Intelligence},
  author       = {{DeepSeek-AI}},
  year         = {2026},
  howpublished = {Technical report, Hugging Face model repository},
  url          = {https://huggingface.co/deepseek-ai/DeepSeek-V4-Pro},
  note         = {Accessed: 2026-05-21}
}

@misc{gan2025ragmcpmitigatingpromptbloat,
      title={RAG-MCP: Mitigating Prompt Bloat in LLM Tool Selection via Retrieval-Augmented Generation}, 
      author={Tiantian Gan and Qiyao Sun},
      year={2025},
      eprint={2505.03275},
      archivePrefix={arXiv},
      primaryClass={cs.AI},
      url={https://arxiv.org/abs/2505.03275}, 
}

@misc{kimiteam2026kimik25visualagentic,
      title={Kimi K2.5: Visual Agentic Intelligence}, 
      author={Kimi Team and Tongtong Bai and Yifan Bai and Yiping Bao and S. H. Cai and Yuan Cao and Y. Charles and H. S. Che and Cheng Chen and Guanduo Chen and Huarong Chen and Jia Chen and Jiahao Chen and Jianlong Chen and Jun Chen and Kefan Chen and Liang Chen and Ruijue Chen and Xinhao Chen and Yanru Chen and Yanxu Chen and Yicun Chen and Yimin Chen and Yingjiang Chen and Yuankun Chen and Yujie Chen and Yutian Chen and Zhirong Chen and Ziwei Chen and Dazhi Cheng and Minghan Chu and Jialei Cui and Jiaqi Deng and Muxi Diao and Hao Ding and Mengfan Dong and Mengnan Dong and Yuxin Dong and Yuhao Dong and Angang Du and Chenzhuang Du and Dikang Du and Lingxiao Du and Yulun Du and Yu Fan and Shengjun Fang and Qiulin Feng and Yichen Feng and Garimugai Fu and Kelin Fu and Hongcheng Gao and Tong Gao and Yuyao Ge and Shangyi Geng and Chengyang Gong and Xiaochen Gong and Zhuoma Gongque and Qizheng Gu and Xinran Gu and Yicheng Gu and Longyu Guan and Yuanying Guo and Xiaoru Hao and Weiran He and Wenyang He and Yunjia He and Chao Hong and Hao Hu and Jiaxi Hu and Yangyang Hu and Zhenxing Hu and Ke Huang and Ruiyuan Huang and Weixiao Huang and Zhiqi Huang and Tao Jiang and Zhejun Jiang and Xinyi Jin and Yu Jing and Guokun Lai and Aidi Li and C. Li and Cheng Li and Fang Li and Guanghe Li and Guanyu Li and Haitao Li and Haoyang Li and Jia Li and Jingwei Li and Junxiong Li and Lincan Li and Mo Li and Weihong Li and Wentao Li and Xinhang Li and Xinhao Li and Yang Li and Yanhao Li and Yiwei Li and Yuxiao Li and Zhaowei Li and Zheming Li and Weilong Liao and Jiawei Lin and Xiaohan Lin and Zhishan Lin and Zichao Lin and Cheng Liu and Chenyu Liu and Hongzhang Liu and Liang Liu and Shaowei Liu and Shudong Liu and Shuran Liu and Tianwei Liu and Tianyu Liu and Weizhou Liu and Xiangyan Liu and Yangyang Liu and Yanming Liu and Yibo Liu and Yuanxin Liu and Yue Liu and Zhengying Liu and Zhongnuo Liu and Enzhe Lu and Haoyu Lu and Zhiyuan Lu and Junyu Luo and Tongxu Luo and Yashuo Luo and Long Ma and Yingwei Ma and Shaoguang Mao and Yuan Mei and Xin Men and Fanqing Meng and Zhiyong Meng and Yibo Miao and Minqing Ni and Kun Ouyang and Siyuan Pan and Bo Pang and Yuchao Qian and Ruoyu Qin and Zeyu Qin and Jiezhong Qiu and Bowen Qu and Zeyu Shang and Youbo Shao and Tianxiao Shen and Zhennan Shen and Juanfeng Shi and Lidong Shi and Shengyuan Shi and Feifan Song and Pengwei Song and Tianhui Song and Xiaoxi Song and Hongjin Su and Jianlin Su and Zhaochen Su and Lin Sui and Jinsong Sun and Junyao Sun and Tongyu Sun and Flood Sung and Yunpeng Tai and Chuning Tang and Heyi Tang and Xiaojuan Tang and Zhengyang Tang and Jiawen Tao and Shiyuan Teng and Chaoran Tian and Pengfei Tian and Ao Wang and Bowen Wang and Chensi Wang and Chuang Wang and Congcong Wang and Dingkun Wang and Dinglu Wang and Dongliang Wang and Feng Wang and Hailong Wang and Haiming Wang and Hengzhi Wang and Huaqing Wang and Hui Wang and Jiahao Wang and Jinhong Wang and Jiuzheng Wang and Kaixin Wang and Linian Wang and Qibin Wang and Shengjie Wang and Shuyi Wang and Si Wang and Wei Wang and Xiaochen Wang and Xinyuan Wang and Yao Wang and Yejie Wang and Yipu Wang and Yiqin Wang and Yucheng Wang and Yuzhi Wang and Zhaoji Wang and Zhaowei Wang and Zhengtao Wang and Zhexu Wang and Zihan Wang and Zizhe Wang and Chu Wei and Ming Wei and Chuan Wen and Zichen Wen and Chengjie Wu and Haoning Wu and Junyan Wu and Rucong Wu and Wenhao Wu and Yuefeng Wu and Yuhao Wu and Yuxin Wu and Zijian Wu and Chenjun Xiao and Jin Xie and Xiaotong Xie and Yuchong Xie and Yifei Xin and Bowei Xing and Boyu Xu and Jianfan Xu and Jing Xu and Jinjing Xu and L. H. Xu and Lin Xu and Suting Xu and Weixin Xu and Xinbo Xu and Xinran Xu and Yangchuan Xu and Yichang Xu and Yuemeng Xu and Zelai Xu and Ziyao Xu and Junjie Yan and Yuzi Yan and Guangyao Yang and Hao Yang and Junwei Yang and Kai Yang and Ningyuan Yang and Ruihan Yang and Xiaofei Yang and Xinlong Yang and Ying Yang and Yi Yang and Yi Yang and Zhen Yang and Zhilin Yang and Zonghan Yang and Haotian Yao and Dan Ye and Wenjie Ye and Zhuorui Ye and Bohong Yin and Chengzhen Yu and Longhui Yu and Tao Yu and Tianxiang Yu and Enming Yuan and Mengjie Yuan and Xiaokun Yuan and Yang Yue and Weihao Zeng and Dunyuan Zha and Haobing Zhan and Dehao Zhang and Hao Zhang and Jin Zhang and Puqi Zhang and Qiao Zhang and Rui Zhang and Xiaobin Zhang and Y. Zhang and Yadong Zhang and Yangkun Zhang and Yichi Zhang and Yizhi Zhang and Yongting Zhang and Yu Zhang and Yushun Zhang and Yutao Zhang and Yutong Zhang and Zheng Zhang and Chenguang Zhao and Feifan Zhao and Jinxiang Zhao and Shuai Zhao and Xiangyu Zhao and Yikai Zhao and Zijia Zhao and Huabin Zheng and Ruihan Zheng and Shaojie Zheng and Tengyang Zheng and Junfeng Zhong and Longguang Zhong and Weiming Zhong and M. Zhou and Runjie Zhou and Xinyu Zhou and Zaida Zhou and Jinguo Zhu and Liya Zhu and Xinhao Zhu and Yuxuan Zhu and Zhen Zhu and Jingze Zhuang and Weiyu Zhuang and Ying Zou and Xinxing Zu},
      year={2026},
      eprint={2602.02276},
      archivePrefix={arXiv},
      primaryClass={cs.CL},
      url={https://arxiv.org/abs/2602.02276}, 
}

@misc{HarnessEngineering,
	author = {OpenAI},
	title = {{H}arness engineering: leveraging {C}odex in an agent-first world},
	howpublished = {\url{https://openai.com/index/harness-engineering/}},
	year = {2026},
	note = {[Accessed 21-05-2026]},
}

@article{xie2024osworld,
  title={Osworld: Benchmarking multimodal agents for open-ended tasks in real computer environments},
  author={Xie, Tianbao and Zhang, Danyang and Chen, Jixuan and Li, Xiaochuan and Zhao, Siheng and Cao, Ruisheng and Hua, Toh J and Cheng, Zhoujun and Shin, Dongchan and Lei, Fangyu and others},
  journal={Advances in Neural Information Processing Systems},
  volume={37},
  pages={52040--52094},
  url={https://dl.acm.org/doi/10.5555/3737916.3739566}, 
  year={2024}
}

@misc{zhang2026clawbenchaiagentscomplete,
      title={ClawBench: Can AI Agents Complete Everyday Online Tasks?}, 
      author={Yuxuan Zhang and Yubo Wang and Yipeng Zhu and Penghui Du and Junwen Miao and Xuan Lu and Wendong Xu and Yunzhuo Hao and Songcheng Cai and Xiaochen Wang and Huaisong Zhang and Xian Wu and Yi Lu and Minyi Lei and Kai Zou and Huifeng Yin and Ping Nie and Liang Chen and Dongfu Jiang and Wenhu Chen and Kelsey R. Allen},
      year={2026},
      eprint={2604.08523},
      archivePrefix={arXiv},
      primaryClass={cs.CL},
      url={https://arxiv.org/abs/2604.08523}, 
}

@article{ding2026wildclawbench,
  title={WildClawBench: A Benchmark for Real-World, Long-Horizon Agent Evaluation},
  author={Ding, Shuangrui and Dai, Xuanlang and Xing, Long and Ding, Shengyuan and Liu, Ziyu and JingYi, Yang and Yang, Penghui and Zhang, Zhixiong and Wei, Xilin and Fang, Xinyu and others},
  journal={arXiv preprint arXiv:2605.10912},
  url={https://arxiv.org/abs/2605.10912},
  year={2026}
}

@article{schick2023toolformer,
  title={Toolformer: Language models can teach themselves to use tools},
  author={Schick, Timo and Dwivedi-Yu, Jane and Dess{\`\i}, Roberto and Raileanu, Roberta and Lomeli, Maria and Hambro, Eric and Zettlemoyer, Luke and Cancedda, Nicola and Scialom, Thomas},
  journal={Advances in neural information processing systems},
  volume={36},
  pages={68539--68551},
  year={2023}
}

@article{patil2024gorilla,
  title={Gorilla: Large language model connected with massive apis},
  author={Patil, Shishir G and Zhang, Tianjun and Wang, Xin and Gonzalez, Joseph E},
  journal={Advances in Neural Information Processing Systems},
  volume={37},
  pages={126544--126565},
  year={2024}
}

@inproceedings{qin2024toolllm,
  title={Toolllm: Facilitating large language models to master 16000+ real-world apis},
  author={Qin, Yujia and Liang, Shihao and Ye, Yining and Zhu, Kunlun and Yan, Lan and Lu, Yaxi and Lin, Yankai and Cong, Xin and Tang, Xiangru and Qian, Bill and others},
  booktitle={International Conference on Learning Representations},
  volume={2024},
  pages={9695--9717},
  year={2024}
}

@inproceedings{react,
  title={ReAct: Synergizing Reasoning and Acting in Language Models},
  author={Yao, Shunyu and Zhao, Jeffrey and Yu, Dian and Du, Nan and Shafran, Izhak and Narasimhan, Karthik and Cao, Yuan},
  booktitle={International Conference on Learning Representations},
  volume={2023},
  year={2023}
}

@article{qian2026toolrl,
  title={Toolrl: Reward is all tool learning needs},
  author={Qian, Cheng and Acikgoz, Emre Can and He, Qi and Wang, Hongru and Chen, Xiusi and Hakkani-Tur, Dilek and Tur, Gokhan and Ji, Heng},
  journal={Advances in Neural Information Processing Systems},
  volume={38},
  pages={105523--105553},
  year={2026}
}

@article{feng2025retool,
  title={Retool: Reinforcement learning for strategic tool use in llms},
  author={Feng, Jiazhan and Huang, Shijue and Qu, Xingwei and Zhang, Ge and Qin, Yujia and Zhong, Baoquan and Jiang, Chengquan and Chi, Jinxin and Zhong, Wanjun},
  journal={arXiv preprint arXiv:2504.11536},
  year={2025}
}

@article{li2025torl,
  title={Torl: Scaling tool-integrated rl},
  author={Li, Xuefeng and Zou, Haoyang and Liu, Pengfei},
  journal={arXiv preprint arXiv:2503.23383},
  year={2025}
}

@article{tool2vec,
  title={Efficient and scalable estimation of tool representations in vector space},
  author={Moon, Suhong and Jha, Siddharth and Erdogan, Lutfi Eren and Kim, Sehoon and Lim, Woosang and Keutzer, Kurt and Gholami, Amir},
  journal={arXiv preprint arXiv:2409.02141},
  year={2024}
}

@article{du2024anytool,
  title={Anytool: Self-reflective, hierarchical agents for large-scale api calls},
  author={Du, Yu and Wei, Fangyun and Zhang, Hongyang},
  journal={arXiv preprint arXiv:2402.04253},
  year={2024}
}

@inproceedings{zheng2024toolrerank,
  title={Toolrerank: Adaptive and hierarchy-aware reranking for tool retrieval},
  author={Zheng, Yuanhang and Li, Peng and Liu, Wei and Liu, Yang and Luan, Jian and Wang, Bin},
  booktitle={Proceedings of the 2024 joint international conference on computational linguistics, language resources and evaluation (LREC-COLING 2024)},
  pages={16263--16273},
  year={2024}
}

@article{smelly,
  title={Model context protocol (mcp) tool descriptions are smelly! towards improving ai agent efficiency with augmented mcp tool descriptions},
  author={Hasan, Mohammed Mehedi and Li, Hao and Rajbahadur, Gopi Krishnan and Adams, Bram and Hassan, Ahmed E},
  journal={arXiv preprint arXiv:2602.14878},
  year={2026}
}

@misc{Steinberger2026openclaw,
	author = {Steinberger, Peter and Koc, Vincent and {Shakker} and Zaidi, Ayaan and Hoffman, Tak and Santana, Gustavo Madeira and {Vignesh} and {github-actions[bot]} and {Shadow} and Avant, Josh and Alexander, Val and Slight, Seb and {Mariano} and Nakazawa, Christoph and {clawsweeper[bot]} and Gutman, Nimrod and Yust, Tyler and Gondhi, Pavan Kumar and Tomlinson, Jacob and {scoootscooob} and Lehman, Josh and {pashpashpash} and Makwana, Neerav and Della-Libera, Gio and Castro, Marcus and Rivera, Agustin and {Sid} and {the sun gif man} and Knight, Alex and Hunt, Harold},
	year = {2026},
	month = {may 23},
	title = {openclaw/openclaw},
	url = {https://github.com/openclaw/openclaw},
	howpublished = {https://github.com/openclaw/openclaw},
}

@inproceedings{BFCL,
  title={The Berkeley Function Calling Leaderboard (BFCL): From Tool Use to Agentic Evaluation of Large Language Models},
  author={Shishir G. Patil and Huanzhi Mao and Fanjia Yan and Charlie Cheng-Jie Ji and Vishnu Suresh and Ion Stoica and Joseph Gonzalez},
  booktitle={International Conference on Machine Learning},
  year={2025},
  url={https://api.semanticscholar.org/CorpusID:283567780}
}

@article{tau-bench,
  title={$\tau$-bench: A Benchmark for Tool-Agent-User Interaction in Real-World Domains},
  author={Shunyu Yao and Noah Shinn and Pedram Razavi and Karthik Narasimhan},
  journal={ArXiv},
  year={2024},
  volume={abs/2406.12045},
  url={https://api.semanticscholar.org/CorpusID:270562578}
}

@article{qwenembedding,
  title={Qwen3 embedding: Advancing text embedding and reranking through foundation models},
  author={Zhang, Yanzhao and Li, Mingxin and Long, Dingkun and Zhang, Xin and Lin, Huan and Yang, Baosong and Xie, Pengjun and Yang, An and Liu, Dayiheng and Lin, Junyang and others},
  journal={arXiv preprint arXiv:2506.05176},
  year={2025}
}

@article{deepseek-v3,
  title={Deepseek-v3. 2: Pushing the frontier of open large language models},
  author={Liu, Aixin and Mei, Aoxue and Lin, Bangcai and Xue, Bing and Wang, Bingxuan and Xu, Bingzheng and Wu, Bochao and Zhang, Bowei and Lin, Chaofan and Dong, Chen and others},
  journal={arXiv preprint arXiv:2512.02556},
  year={2025}
}

\appendix
\section{Additional Details}
\label{app:additional details}

\subsection{Failure Analysis Taxonomy}
\label{app:failure_analysis_taxonomy}

Each failed trajectory is assigned exactly one primary error category, corresponding to the earliest root cause that prevents task completion. Downstream symptoms triggered by that root cause are not relabeled.

\begin{enumerate}
    \item[\textbf{(1)}] \textbf{Query Formulation Error.}
    The agent fails to express the correct tool need at the planning or decomposition stage, before discovery can help.
    The diagnostic signal is that the queries in \texttt{search\_details} do not cover one or more required goals of the task, such as missing an implicit upstream step, ignoring a temporal or unit constraint, or producing a query that is too broad or too narrow to match any usable tool.
    For the default setting where no discovery step exists, this category is not applicable.

    \item[\textbf{(2)}] \textbf{Discovery Error.}
    discovery errors occur when the agent issues a semantically valid tool request, 
    but the discovery module fails to return any tool with the capability required for the corresponding task.
    This category reflects a capability mismatch introduced by the discovery stage, 
    rather than a strict mismatch with the annotated ground-truth tool.
    When the retrieved set contains a functionally equivalent tool, the failure is 
    attributed to downstream tool use instead of discovery.
    This category is not applicable to the default setting, where tools are directly exposed without discovery.

    \item[\textbf{(3)}] \textbf{Tool-Use Error.}
    A required server or tool is available to the agent, either retrieved or exposed by the default setting, and the agent calls it but uses it incorrectly.
    Sub types include wrong arguments, schema mismatches, malformed JSON, wrong call order, missing prerequisite calls, misinterpretation of tool output, and selecting a distractor over a present correct tool.

    \item[\textbf{(4)}] \textbf{Reasoning / Other Error.}
    All required tools are available and were used correctly, but the task still fails for reasons that are not discovery  or tool-use-related.
    Sub types include incomplete final reasoning, such as missing a claim that was actually retrieved during execution; premature stopping or self misjudged completion; forgetting an earlier user constraint at answer time; and external execution failures such as timeouts, server unavailability, step limits, context limits, or sandbox errors.
\end{enumerate}

\subsection{Server Discovery with Different Decomposition Models}
\label{app:decompose_models}

Table~\ref{tab:decompose_model_benchmark} further studies the influence of decomposition models on server discovery. 
Overall, DeepSeek-V4-Pro achieves the strongest performance, especially under the global setting.
One possible reason is that the tool-side intentions in our graph are generated by DeepSeek-V3.2.
Therefore, decomposition models from the same DeepSeek family may produce subtasks and query-side intentions with a more similar linguistic style and semantic distribution.
This upstream--downstream consistency can make the generated discovery queries better aligned with the intention nodes in the graph, leading to more accurate server discovery.
In contrast, models from different families may express the same functional need with different wording or abstraction levels, which can slightly weaken intention matching even when the decomposition itself is reasonable.

\begin{table}[t]
\centering
\scriptsize
\setlength{\tabcolsep}{3pt}
\renewcommand{\arraystretch}{0.96}
\caption{Server discovery performance on each benchmark with different decomposition models.}
\label{tab:decompose_model_benchmark}
\resizebox{\linewidth}{!}{%
\begin{tabular}{llccc}
\toprule
\textbf{Decompose Model} & \textbf{Benchmark} & \textbf{Recall@5} & \shortstack{\textbf{Full-Recall@5}} & \textbf{MRR} \\
\midrule
\multicolumn{5}{l}{\textbf{Restricted setting}} \\
\addlinespace

DeepSeek-V3.2   & MCPBench     & 0.9744 & 0.9327 & \textbf{1.0000} \\
                & MCP-Universe & 0.9310 & 0.8664 & 0.9112 \\
                & MCP-Atlas    & 0.5777 & 0.2460 & 0.7339 \\
\cmidrule(lr){2-5}
                & \textbf{Average} & 0.7251 & 0.5036 & 0.8162 \\
\addlinespace

DeepSeek-V4-Pro & MCPBench     & \textbf{0.9936} & \textbf{0.9810} & \textbf{1.0000} \\
                & MCP-Universe & 0.9116 & 0.8230 & 0.9348 \\
                & MCP-Atlas    & \textbf{0.5897} & \textbf{0.2680} & \textbf{0.7456} \\
\cmidrule(lr){2-5}
                & \textbf{Average} & 0.7293 & 0.5110 & 0.8298 \\
\addlinespace

Kimi-K2.5       & MCPBench     & 0.9776 & 0.9420 & 0.9952 \\
                & MCP-Universe & \textbf{0.9397} & \textbf{0.8790} & 0.9340 \\
                & MCP-Atlas    & 0.5839 & 0.2520 & 0.7371 \\
\cmidrule(lr){2-5}
                & \textbf{Average} & \textbf{0.7316} & \textbf{0.5120} & 0.8238 \\
\addlinespace

Kimi-K2.6       & MCPBench     & 0.9840 & 0.9520 & \textbf{1.0000} \\
                & MCP-Universe & 0.9181 & 0.8410 & \textbf{0.9445} \\
                & MCP-Atlas    & 0.5896 & 0.2580 & 0.7439 \\
\cmidrule(lr){2-5}
                & \textbf{Average} & 0.7298 & 0.5060 & \textbf{0.8314} \\

\midrule
\multicolumn{5}{l}{\textbf{Global setting}} \\
\addlinespace

DeepSeek-V3.2   & MCPBench     & 0.8189 & 0.6827 & 0.8550 \\
                & MCP-Universe & 0.4547 & 0.3750 & 0.4056 \\
                & MCP-Atlas    & 0.2881 & 0.0420 & 0.3823 \\
\cmidrule(lr){2-5}
                & \textbf{Average} & 0.4004 & 0.2140 & 0.4476 \\
\addlinespace

DeepSeek-V4-Pro & MCPBench     & 0.8974 & 0.7600 & 0.9199 \\
                & MCP-Universe & \textbf{0.5438} & 0.3750 & \textbf{0.5089} \\
                & MCP-Atlas    & \textbf{0.3172} & 0.0540 & \textbf{0.4262} \\
\cmidrule(lr){2-5}
                & \textbf{Average} & \textbf{0.4523} & 0.2310 & \textbf{0.5105} \\
\addlinespace

Kimi-K2.5       & MCPBench     & \textbf{0.9006} & \textbf{0.7880} & \textbf{0.9258} \\
                & MCP-Universe & 0.5244 & 0.3660 & 0.5057 \\
                & MCP-Atlas    & 0.3021 & \textbf{0.0580} & 0.3930 \\
\cmidrule(lr){2-5}
                & \textbf{Average} & 0.4383 & \textbf{0.2340} & 0.4906 \\
\addlinespace

Kimi-K2.6       & MCPBench     & 0.8846 & 0.7690 & 0.8926 \\
                & MCP-Universe & 0.5309 & \textbf{0.3790} & 0.4841 \\
                & MCP-Atlas    & 0.3092 & 0.0520 & 0.4108 \\
\cmidrule(lr){2-5}
                & \textbf{Average} & 0.4423 & 0.2320 & 0.4911 \\

\bottomrule
\end{tabular}
}
\end{table}

\subsection{Extra Details of Server Collection}
\label{app:server_collection_details}

\begin{figure}[t]
    \centering
    \includegraphics[width=\linewidth]{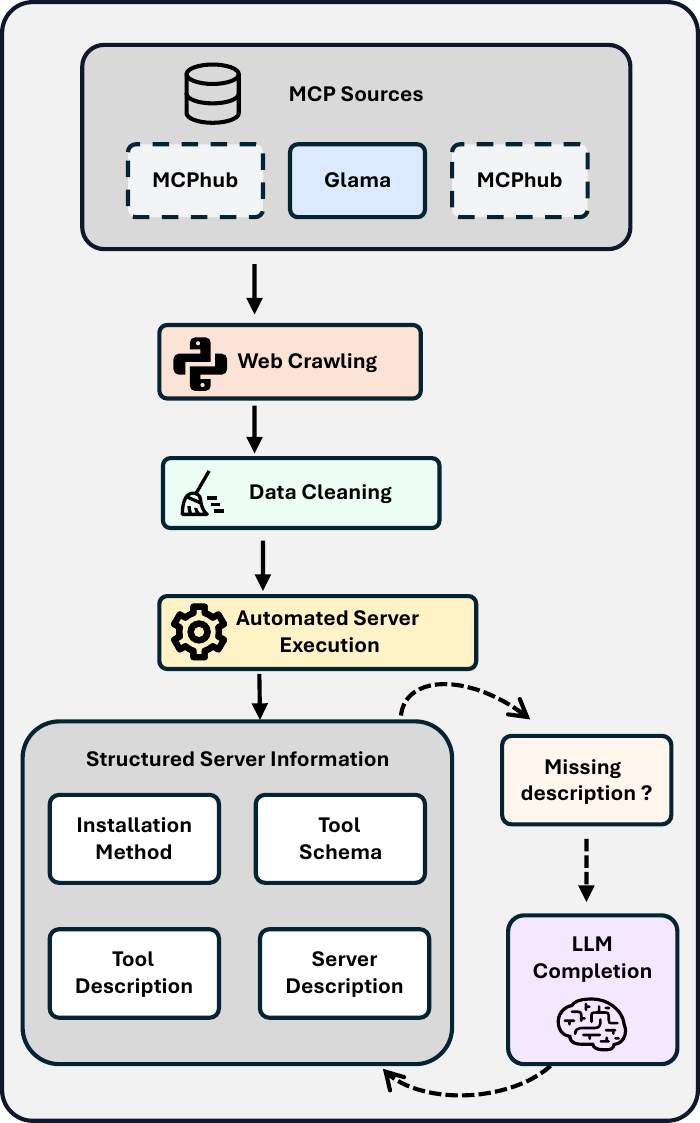}
    \caption{Overview of our MCP server collection and metadata extraction pipeline.}
    \label{fig:server_collection_pipeline}
\end{figure}

We first crawled MCP server metadata from MCPhub, Glama, and Smithery.
Because these marketplaces changed frequently and were sometimes unavailable, we use a fixed Glama snapshot as the final source to reduce platform-induced noise.
We then clean the noisy metadata and apply an execution-based extraction pipeline: extracting launch configurations, deploying servers when possible, querying successfully launched servers through the MCP interface, and collecting their tool schemas.
For missing server or tool descriptions, we use an LLM to complete the metadata from available server information, tool names, and schemas.
This yields a structured server--tool library for graph construction and discovery evaluation, as shown in Figure~\ref{fig:server_collection_pipeline}.

To distinguish between reported collection scale and the scale actually used in evaluation, we compare SING with recent MCP-related studies, including MCPToolBench++~\citep{fan2025mcptoolbenchlargescaleai}, MCPAgentBench~\citep{liu2026mcpagentbenchrealworldtaskbenchmark}, MCP-Flow~\citep{wang2026mcpflowfacilitatingllmagents}, LiveMCPBench~\citep{mo2026livemcpbenchagentsnavigateocean}, MCP-Zero~\citep{fei2025mcpzeroactivetooldiscovery}, DIVE~\citep{chen2026divescalingdiversityagentic}, and TOUCAN~\citep{xu2025toucansynthesizing15mtoolagentic}. Table~\ref{tab:mcp_collection_comparison} shows that SING evaluates the largest explicitly reported unique experimental server--tool pool among these works.

\begin{table}[t]
\centering
\scriptsize
\setlength{\tabcolsep}{2.0pt}
\renewcommand{\arraystretch}{1.18}
\caption{Comparison of MCP-related studies by collected and actually used scale. S/T denotes servers/tools; N/S denotes not specified.}
\label{tab:mcp_collection_comparison}
\begin{tabularx}{\columnwidth}{@{}p{0.24\columnwidth} p{0.22\columnwidth} p{0.16\columnwidth} X@{}}
\toprule
\textbf{Study} &
\multicolumn{2}{c}{\textbf{Collected}} &
\textbf{Experimented / Used} \\
\cmidrule(lr){2-3}
&
\textbf{Numbers} &
\textbf{Domains} &
\textbf{Numbers} \\
\midrule

\makecell[l]{MCPTool\\Bench++} &
$>$4K S &
12 domains &
\makecell[l]{12 S / 87 T;\\1,509 samples} \\
\addlinespace[1.5pt]

\makecell[l]{MCPAgent\\Bench} &
\makecell[l]{9,714 S /\\$>$20K T} &
2 general &
\makecell[l]{N/S unique S/T;\\$\sim$4K T exp.} \\
\addlinespace[1.5pt]

MCP-Flow &
\makecell[l]{1,166 S /\\11,536 T} &
10 domains &
\makecell[l]{356 S;\\6,439 traj.} \\
\addlinespace[1.5pt]

LiveMCPBench &
5,588 configs &
6 domains &
70 S / 527 T \\
\addlinespace[1.5pt]

MCP-Zero &
\makecell[l]{308 S /\\2,797 T} &
N/S &
\makecell[l]{308 S /\\2,797 T} \\
\addlinespace[1.5pt]

DIVE &
373 T &
\makecell[l]{2 general +\\4 expert} &
N/S exact S/T \\
\addlinespace[1.5pt]

TOUCAN &
$\sim$2,800 specs &
N/S &
\makecell[l]{495 S /\\$>$2,000 T} \\
\addlinespace[1.5pt]

SING (Ours) &
\makecell[l]{1,141 S /\\21,618 T} &
\makecell[l]{8 major /\\15 sub.} &
\makecell[l]{\textbf{779 S} /\\\textbf{7,471 T}} \\

\bottomrule
\end{tabularx}
\end{table}

\section{Prompts}
\subsection{Single-Tool Query Generation Prompt}
We use the following prompt to synthesize realistic user instructions and corresponding function-call arguments for each individual MCP tool.

\begin{lstlisting}[style=promptstyle,language=Python]
GENERATE_PROMPT = """You are generating realistic user instructions for an MCP (Model Context Protocol) tool.

Given the tool information below, generate {n_queries} diverse user instructions that would naturally invoke this tool.
For each instruction, also provide the corresponding function call arguments.

Tool information:
- Server: {server_name}
- Server description: {server_desc}
- Tool name: {tool_name}
- Tool description: {tool_desc}
{param_section}

Rules:
1. Each instruction should be a natural user request (as if a user is asking an AI assistant)
2. Instructions should be DIVERSE -- vary the specific entities, parameters, and complexity
3. Arguments must match the tool's parameter schema -- use realistic values
4. Do NOT use placeholder values like "xxx" or "example" -- use realistic but fictional data
5. Output ONLY a JSON array, each element has "source_instruction" (string) and "arguments" (object)

Example output format:
[
  {{"source_instruction": "Find all flights from New York to London on December 15th", "arguments": {{"origin": "JFK", "destination": "LHR", "date": "2024-12-15"}}}},
  {{"source_instruction": "Search for the cheapest flights to Tokyo next week", "arguments": {{"destination": "NRT", "date": "2024-12-20", "sort": "price"}}}}
]

Generate {n_queries} diverse instructions:"""
\end{lstlisting}

\subsection{Multi-Tool Query Extension Prompt}
We use the following prompt to extend an existing user query with an additional MCP tool from the same server when the extension forms a natural workflow.

\begin{lstlisting}[style=promptstyle,language=Python]
EXTEND_PROMPT = """You are extending a user's query to naturally incorporate an additional MCP tool from the same server.

Server: {server_name}
Server description: {server_desc}

Current user query: "{current_query}"

Tools already being used:
{used_tools_section}

Available additional tools (NOT yet used):
{available_tools_section}

Your task:
1. Evaluate whether any of the available tools can NATURALLY extend the user's query
2. The extension must be a logical follow-up or complement -- NOT forced or artificial
3. The extended query should read as a SINGLE coherent user request that naturally requires both old and new tools

Key criteria for a MEANINGFUL extension:
- The new tool provides genuine additional value (e.g., transforming results, enriching data, performing a next step)
- A real user would plausibly combine these tools in one request
- It is NOT just restating the same question, repeating similar functionality, or doing something unrelated
- The tools should have a clear logical dependency or workflow connection

If NO tool can meaningfully extend the query, return:
{{"can_extend": false, "reason": "brief explanation"}}

If extension IS meaningful, return:
{{"can_extend": true, "extended_query": "the full extended user request combining all tools", "new_tool": "exact_tool_name_from_available_list", "new_arguments": {{}}, "reason": "why this extension is natural"}}

Output ONLY valid JSON:"""
\end{lstlisting}

\subsection{Batch Intention Generation Prompt}
We use the following prompt to generate tool-action intentions from synthesized queries.

\begin{lstlisting}[style=promptstyle,language=Python]
BATCH_INTENTION_PROMPT = """You are analyzing user queries that invoke MCP tools to generate tool-action intentions.

For EACH tool, based on its sample queries, generate 3-4 intention phrases. Each intention MUST follow this format:
  "[verb] [specific data object/resource]"

Rules:
1. Each intention should describe a CONCRETE tool action, e.g. "fetch historical stock prices", "geocode address to coordinates", "parse HTML table into structured data"
2. Include 1-2 intentions for IMPLICIT prerequisite or follow-up actions -- tools the user would ALSO need in the same workflow but didn't explicitly mention
3. Do NOT describe user scenarios or personas -- no "A researcher wants to..." or "planning a trip..."
4. Do NOT rephrase the tool description -- focus on WHAT DATA the tool operates on and HOW
5. Keep each intention 4-8 words, starting with an action verb

Output ONLY a JSON object mapping tool_key to intention array, no explanation.

Example: {{"ServerA__geocode": ["resolve city names to GPS coordinates", "convert postal codes to lat/lng", "prepare location input for route planning", "validate address format before geocoding"]}}

Tools to analyze:
{tool_entries}"""
\end{lstlisting}

\section{Examples}
\subsection{Query and Intention Synthesis Example}
\label{app:query_intention_synthesis_example}

Table~\ref{tab:query_intention_chain_example} shows an  example of our query and intention synthesis process.
Starting from a single GitHub server, we synthesize a natural language query and expand it into a multi-tool chain.
For each tool in the chain, we extract atomic intentions that describe reusable user goals supported by the tool, rather than directly paraphrasing the tool description.

\begin{table}[ht]
\centering
\footnotesize
\setlength{\tabcolsep}{3pt}
\renewcommand{\arraystretch}{1.12}
\begin{tabularx}{\columnwidth}{@{}p{0.13\columnwidth}X@{}}
\toprule
\textbf{Item} & \textbf{Example} \\
\midrule
Server &
\texttt{github} \\

Query &
Search Rust machine learning repositories, inspect repository code files, and search for specific implementation patterns. \\

Chain &
\texttt{search\_repositories} $\rightarrow$
\texttt{github\_get\_code} $\rightarrow$
\texttt{search\_code} \\

\midrule
Step 1 &
\textbf{Tool:} \texttt{search\_repositories}. 
\textbf{Desc.:} Searches for repositories on GitHub. 
\textbf{I/O:} repository search query $\rightarrow$ repository search results. 
\textit{Intentions:} filter projects by creation date; rank repositories by fork count; validate repository metadata before cloning. \\

Step 2 &
\textbf{Tool:} \texttt{github\_get\_code}. 
\textbf{Desc.:} Retrieves code files from a GitHub repository. 
\textbf{I/O:} repository URL and file filters $\rightarrow$ repository code files. 
\textit{Intentions:} extract specific function implementations; filter code files by path patterns; retrieve code metadata and structure. \\

Step 3 &
\textbf{Tool:} \texttt{search\_code}.
\textbf{Desc.:} Searches code across GitHub repositories.
\textbf{I/O:} code search query $\rightarrow$ code search results.
\textit{Intentions:} validate code patterns before analysis; locate specific function implementations. \\
\bottomrule
\end{tabularx}
\caption{Example of query and intention synthesis for one GitHub tool chain.}
\label{tab:query_intention_chain_example}

\end{table}

\subsection{Tool Substitutability Example on MCP-Atlas}
\label{app:atlas_substitution_case}

Table~\ref{app:atlas_substitution_case} illustrates that the official gold trajectory is not always the only valid solution path. In the global MCP ecosystem, different servers may expose functionally similar tools, such as geocoding, web search, or local computation tools. 

\begin{table}[ht]
\centering
\scriptsize
\renewcommand{\arraystretch}{1.15}
\setlength{\tabcolsep}{4pt}
\caption{A representative MCP-Atlas execution under the 779-server global pool (task \texttt{689cd6f8\ldots fee}). Both gold-trajectory servers (\texttt{national parks}, \texttt{mcp\_code\_executor}) are bypassed: SING returns five alternative geocoders, the agent fails over from Google Maps to OpenStreetMap, and computes the Haversine distance with the LLM itself. Coverage score: \textbf{1.0} (1/1 claim fully covered).}
\label{tab:atlas_substitute_case}
\begin{tabular}{@{}p{0.04\columnwidth} p{0.90\columnwidth}@{}}
\toprule
\multicolumn{2}{@{}p{0.94\columnwidth}@{}}{%
\textbf{Prompt:} \textit{``Calculate the straight-line distance in km between Gates of the Arctic National Park and Glacier Bay National Park.''} \newline
\textbf{Gold servers:} \texttt{national parks}, \texttt{mcp\_code\_executor}\quad(\textbf{neither used})} \\
\midrule
1 & \texttt{search\_tools(query="geocode national park lat lon", top\_k=5)} \newline
    $\rightarrow$ \textit{5 servers from 779-pool:} \texttt{mcp\_google\_maps}, \texttt{amap}, \texttt{mcp-google-maps-alt}, \texttt{playwright}, \texttt{open-streetmap-mcp} \\
\addlinespace[2pt]
2 & \texttt{get\_tool\_schema(mcp\_google\_maps.get\_geocode)} \newline
    $\rightarrow$ \texttt{tools: []} \textit{(server unavailable)} \\
\addlinespace[2pt]
3 & \texttt{get\_tool\_schema(open-streetmap-mcp.geocode\_address)} \newline
    $\rightarrow$ \textit{schema OK; takes} \texttt{address: str}\textit{, returns} \texttt{lat, lon, \ldots} \\
\addlinespace[2pt]
4 & \texttt{execute\_tool(open-streetmap-mcp.geocode\_address,} \newline
    \quad\texttt{addr="Gates of the Arctic National Park")} \newline
    $\rightarrow$ \texttt{lat=66.8661, lon=-155.1352} \\
\addlinespace[2pt]
5 & \texttt{execute\_tool(open-streetmap-mcp.geocode\_address,} \newline
    \quad\texttt{addr="Glacier Bay National Park")} \newline
    $\rightarrow$ \texttt{lat=58.8142, lon=-136.8721} \\
\addlinespace[2pt]
-- & \textit{LLM Haversine (no tool call)} \newline
    \textit{$\rightarrow$ $\approx 1{,}279$ km \quad (claim: $1{,}283.5$ km, judged fulfilled)} \\
\bottomrule
\end{tabular}
\end{table}

\subsection{Error Examples by Type}
\label{app:error_examples}
Table~\ref{tab:error_examples} presents representative failures from our MCP-Atlas error analysis. These examples clarify the distinction between query formulation, discovery, tool use, and reasoning failures.

\begin{table}[ht]
\centering
\footnotesize
\setlength{\tabcolsep}{4pt}
\renewcommand{\arraystretch}{1.05}
\caption{Representative failure examples for each error category in the MCP-Atlas error analysis. Each row shows the user task, the relevant discovery / tool-use evidence, and the root cause of failure.}
\label{tab:error_examples}
\begin{tabularx}{0.99\columnwidth}{@{}p{1.6cm} >{\raggedright\arraybackslash}X@{}}
\toprule
\textbf{Category} & \textbf{Example failure trajectory} \\
\midrule
\textbf{\scriptsize Query}\newline\textbf{\scriptsize Formulation} &
\textbf{Task.} ``What was the moon phase the day when the author of \emph{The Haunting of Hill House} died, but in the book's publication year?''\newline
\textbf{Issued queries.} Only variants of ``moon phase historical date'' and a generic ``wikipedia knowledge search''.\newline
\textbf{Failure.} The decomposition skips the implicit upstream sub-goal of identifying the author and her death date (Shirley Jackson, 1965); no query ever targets a biography-lookup tool, so discovery is never given a chance to surface a usable tool for the missing sub-goal. \\
\midrule
\textbf{Discovery} &
\textbf{Task.} ``From the cloud NoSQL video-game-store database, how many Twitter posts were made in the year Kobe Bryant passed away?''\newline
\textbf{Issued query.} ``cloud nosql video game store database query twitter posts'' (semantically appropriate).\newline
\textbf{Retrieved.} \toolid{mongodb::list-databases}, \toolid{mongodb::db-stats}, \toolid{fetch::fetch}, \ldots\newline
\textbf{Failure.} The required data lives in a Cassandra / Astra DB, but discovery consistently returns only MongoDB- and fetch-style tools across more than fifteen reformulations. No functionally equivalent NoSQL client is ever returned, so the agent cannot reach the social-media collection regardless of how it calls the available tools. \\
\midrule
\textbf{Tool Use} &
\textbf{Task.} ``Translate the message of the first commit made on 2025-07-25 in the repo \emph{A2A} to Spanish.''\newline
\textbf{Available.} \toolid{github::list_commits}, \toolid{git::git_log}, \ldots\ (the correct capability is retrieved).\newline
\textbf{Failure.} The agent calls \toolid{github::list_commits} with \texttt{owner="a2aproject"} instead of the correct \texttt{owner="google"}, retrieving commits from an unrelated fork; it later translates the wrong commit. The tool was the right one and was actually invoked---only the parameter binding was wrong. \\
\midrule
\textbf{Reasoning / Other} &
\textbf{Task.} ``What is the credit line of the Met's painting \emph{Aristotle with a Bust of Homer}, and what was the full name of the mother of its artist?''\newline
\textbf{Trace.} \toolid{metmuseum-mcp::search-museum-objects} $\rightarrow$ \toolid{get-museum-object} $\rightarrow$ \toolid{open-library::get_authors_by_name}; the agent correctly retrieves both the credit line and Rembrandt's mother's full name.\newline
\textbf{Failure.} The final answer omits an explicit claim that Rembrandt is the artist of the painting, which is required by the judge. All discovery and tool calls are correct; the failure is in answer construction---a retrieved fact that is never stated in the response. \\
\bottomrule
\end{tabularx}
\end{table}

\section{Benchmark Comparison}
\label{appendix:benchmark_details}
This section summarizes the downstream task benchmarks used in our evaluation.
Table~\ref{tab:benchmark_details} provides a compact comparison of MCP-Universe, MCP-Atlas, and MCP-Bench in terms of task scale, server/tool settings, task style, cross-server requirements, and evaluation protocols.

\begin{table*}[p]
\centering
\small
\renewcommand{\arraystretch}{1.25}
\caption{Comparison of the three downstream task benchmarks used in our evaluation. We list only dataset-level properties; our execution statistics are reported separately in the main results.}
\label{tab:benchmark_details}
\begin{tabular}{@{}p{0.17\textwidth} p{0.24\textwidth} p{0.24\textwidth} p{0.24\textwidth}@{}}
\toprule
\textbf{Property} & \textbf{MCP-Universe}~\cite{mcpuniverse} & \textbf{MCP-Atlas}~\cite{bandi2026mcpatlaslargescalebenchmarktooluse} & \textbf{MCP-Bench}~\cite{wang2025mcpbenchbenchmarkingtoolusingllm} \\
\midrule
\# Tasks & 206 & 500 & 104 \\
\midrule
Server pool & 1--3 servers per task, scoped by domain & 36 real-world MCP servers ($\sim$307 tools) & 28 fixed MCP servers ($\sim$250 tools) \\
\midrule
GT servers per task & 1--3, fixed by domain & 2--4, derived from gold trajectory & 1 (Single), 2 (Multi-2), 3 (Multi-3) + 10 distractor servers \\
\midrule
Task domains & 7 explicit domains: financial analysis, location \& navigation, repository management, browser automation, web search, 3D design, multi-server & No official category; broadly cross-domain (real estate, healthcare, finance, travel, entertainment, code, etc.) & No formal category; tasks span biomedical, scientific, financial, geographic, cultural, and developer-tooling servers \\
\midrule
Task generation & Human-curated & Human-curated narratives with annotated gold trajectories & LLM-synthesized with dual-track design \\
\midrule
Task style & Role-played but procedurally precise instructions with explicit numerical/format constraints & Long colloquial narratives encoding \emph{implicit} temporal and factual constraints (e.g.\ ``the month Fred White passed away'') & Paired \textit{fuzzy} user description (visible to agent) + \textit{concrete} description and dependency graph (visible only to judge) \\
\midrule
Cross-server requirement & Optional in most domains; explicit in the multi-server split (25 tasks) & Almost always cross-server (2--4 servers per task) & Mandatory by design for Multi-2/Multi-3; Single tasks still issue many calls within one server \\
\midrule
Evaluation protocol & Per-task programmatic evaluators (rule-based, with LLM-as-judge for the web-search domain) & GTFA fact-claim coverage, judged by an LLM & Six-dimension LLM judge (fulfillment, grounding, tool appropriateness, parameter accuracy, dependency awareness, parallelism) combined with rule-based schema metrics \\
\midrule
Native metrics & Success Rate (all evaluators must pass) & Pass Rate ($\textit{coverage} \geq 0.75$) and Mean Coverage & Overall Score $\in [0,1]$, aggregated from schema understanding, task completion, tool selection, and planning effectiveness \\
\bottomrule
\end{tabular}
\end{table*}

% TODO: Add appendix content here.
\end{document}